\documentclass{ecai} 

\usepackage{latexsym}
\usepackage{amssymb}
\usepackage{amsmath}
\usepackage{amsthm}
\usepackage{booktabs}
\usepackage{enumitem}
\usepackage{graphicx}
\usepackage{color}
\usepackage{subfigure}
\usepackage{multirow}
\usepackage{multicol}
\usepackage{algorithm}
\usepackage{algorithmic}
\usepackage{setspace}
\definecolor{softgreen}{RGB}{7,188,59} 

\newcommand{\BibTeX}{B\kern-.05em{\sc i\kern-.025em b}\kern-.08em\TeX}

\begin{document}


\begin{frontmatter}


\paperid{892} 


\title{Decoupled Competitive Framework for Semi-supervised Medical Image Segmentation}


\author[A]{\fnms{Jiahe}~\snm{Chen}\footnote{Equal contribution.}}
\author[A]{\fnms{Jiahe}~\snm{Ying}\footnotemark}
\author[A]{\fnms{Shen}~\snm{Wang}} 
\author[A]{\fnms{Jianwei}~\snm{Zheng}\thanks{Corresponding Author. Email: zjw@zjut.edu.cn}} 

\address[A]{Zhejiang University of Technology}


\begin{abstract}
Confronting the critical challenge of insufficiently annotated samples in medical domain, semi-supervised medical image segmentation (SSMIS) emerges as a promising solution. Specifically, most methodologies following the Mean Teacher (MT) or Dual Students (DS) architecture have achieved commendable results. However, to date, these approaches face a performance bottleneck due to two inherent limitations, \textit{e.g.}, the over-coupling problem within MT structure owing to the employment of exponential moving average (EMA) mechanism, as well as the severe cognitive bias between two students of DS structure, both of which potentially lead to reduced efficacy, or even model collapse eventually. To mitigate these issues, a Decoupled Competitive Framework (DCF) is elaborated in this work, which utilizes a straightforward competition mechanism for the update of EMA, effectively decoupling students and teachers in a dynamical manner. In addition, the seamless exchange of invaluable and precise insights is facilitated among students, guaranteeing a better learning paradigm. The DCF introduced undergoes rigorous validation on three publicly accessible datasets, which encompass both 2D and 3D datasets. The results demonstrate the superiority of our method over previous cutting-edge competitors. Code will be available at \url{https://github.com/JiaheChen2002/DCF}.
\end{abstract}

\end{frontmatter}


\section{Introduction}
Medical image segmentation (MIS) is critical in modern healthcare, offering clinicians vital information to monitor disease progression and develop treatment strategies. The advent of neural networks, especially supervised deep learning methods, has significantly advanced this field, leading to unparalleled performance in various segmentation tasks \citep{ronneberger2015u,zhou2019unetplusplus, wang2018deepigeos,MISSFormer,yang2023directional}. However, the practical model efficacy largely depends on the availability of massive laboriously annotated datasets, which not only demands specialized knowledge but is also extremely costly and time-consuming \citep{Grnberg2017AnnotatingMI,taleb20203d}.
To save considerable resources, recent years have witnessed an increasing focus on exploring methods like semi-supervised learning (SSL) to reduce the annotation burden in MIS domain \citep{chen2023magicnet,Bai2023BidirectionalCF,Miao2023CauSSLCS}.

\begin{figure}[h]
\centering
\includegraphics[width=0.95\linewidth]{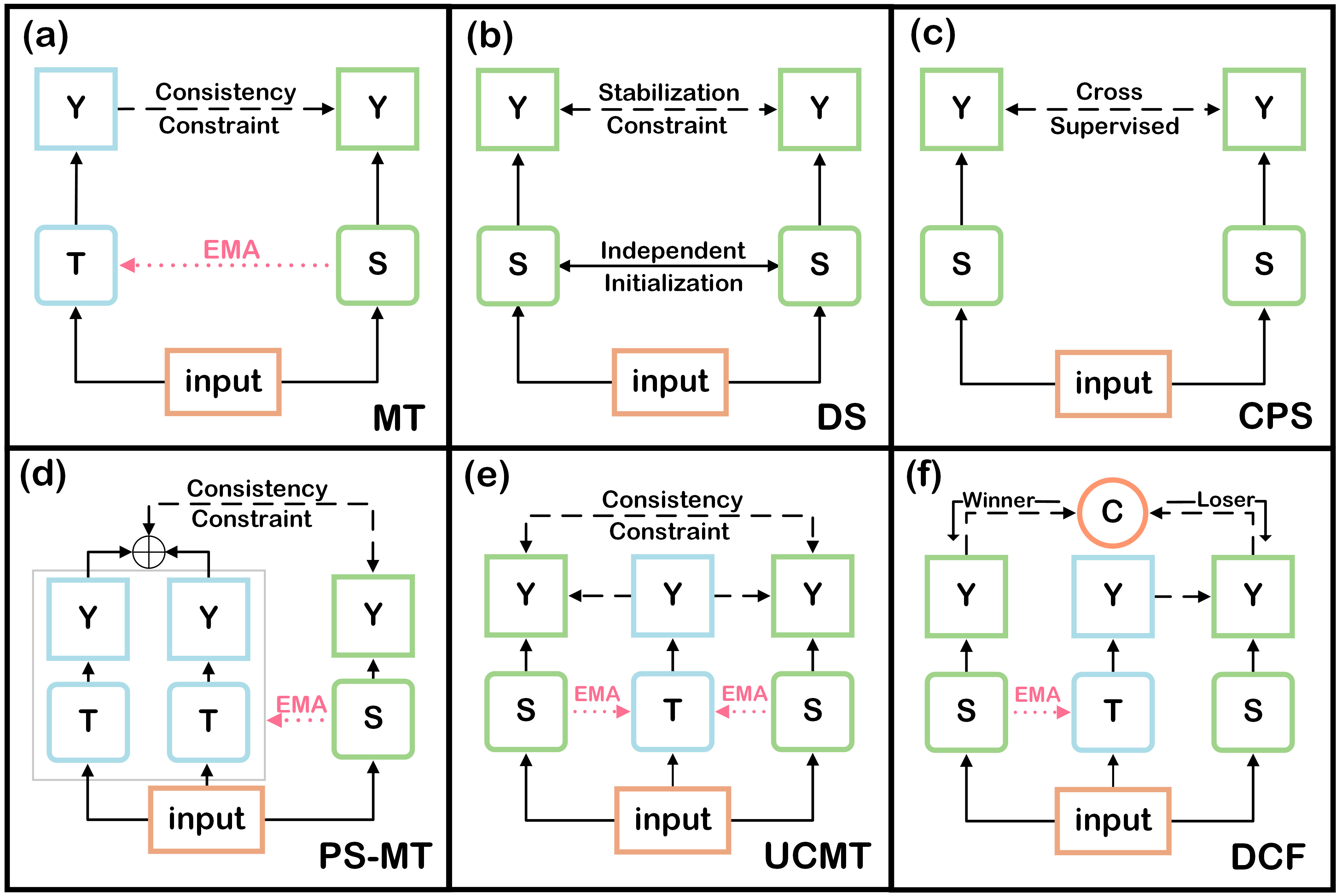}
\caption{Different SSL structures. (a) Mean Teacher. (b) Dual Student. (c) Cross Pseudo Supervision. (d) Perturbed and Strict Mean Teachers. (e) Uncertainty-guided Collaborative Mean Teacher. (f) Decoupled Competitive Framework (ours).}
\label{figure1}
\end{figure}

The advantage of semi-supervised image segmentation lies particularly in leveraging unlabeled data to favor better segmentation. Within this framework, two key strategies, \textit{i.e.}, pseudo-label supervision \citep{zhang2021flexmatch,kwon2022semi,cascante2021curriculum} and consistency regularization \citep{ouali2020semi,liu2022perturbed,mittal2019semi}, have been intensively investigated. Meanwhile, co-training or mutual learning paradigm,  which can be regarded as a combination of the above two methods, has achieved promising results \citep{Shen2023CotrainingWH, xia2020uncertainty,chen2021semi}. Typically, the Mean Teacher model \citep{tarvainen2017mean} has dominated in this field for a long time, inspiring numerous followers seeking notable advancements \citep{yu2018pu, liu2022perturbed, Shen2023CotrainingWH, Fussnet}. Yet, Ke et al. \citep{ke2019dual} and Zhao et al. \citep{10120761} have demonstrated that MT-based methods encounter performance limitations arising from the treatment of exponential moving average (EMA). Besides, employing one coupled EMA teacher is insufficient to adequately support the student model. To break the performance bottleneck, it is often necessary to integrate additional modules with an intricate architecture into MT-based methodologies, such as MagicNet \citep{chen2023magicnet} and BCP \citep{Bai2023BidirectionalCF}. 

The teacher-less architecture is yet another research branch, which offers a promising avenue to mitigate the issue of overly tight coupling. Without a teacher model, the main challenge lies in facilitating efficient knowledge extraction and exchange between two autonomous models. Ke et al. \citep{ke2019dual} introduced Dual Student (DS) as a remedy, replacing the teacher with another student, and integrating a stabilization constraint during training stage. Following this line, Zhao et al. \citep{10120761} additionally integrated region-level uncertainty estimation to ensure better performance.

While the potential of DS is widely acknowledged, successful practices again require the integration of intricate constraints within student models; otherwise, it may lead to model collapse due to the abnormal exchange of erroneous information, which remains as an emergent issue to be solved. Upon this aspiration, we propose a straightforward yet potent solution, \textit{i.e.}, a Decoupled Competitive Framework (DCF), whose disparity against current architecture is given in Figure 1. Note that the real-time performance of two student models can be iteratively assessed. On that basis, the superior-performing student model shall be leaned to update the EMA of teacher model. The teacher model then further provides pseudo-labels to the inferior-performing student to favor an improvement. Through this dynamic mechanism, both student models have the opportunity to contribute to EMA update of the teacher model, which naturally reduces the coupling between a single teacher-student pair. Technically, our work makes three primary contributions.

\begin{itemize}
\item With the deficiencies of MT-based and DS-based methods deeply scrutinized, we engineer a novel Decoupled Competitive Framework, effectively surmounting the bottleneck of existing models.
\item An efficient competition and mentoring mechanism is crafted to mitigate the tight coupling between the teacher's parameters and the individual students, thereby augmenting students' capacity to acquire valuable knowledge.
\item Yet without any sophisticated modules employed, our DCF sets new state-of-the-art scores among three benchmark datasets, namely left atrium segmentation in MRI, pancreas segmentation in CT scans, and dermoscopy images.
\end{itemize}


\vspace{-10pt}
\section{Related Work}

\subsection{Semi-supervised Learning}
\vspace{-2pt}
Semi-supervised learning (SSL) is a widely employed method in numerous computer vision tasks \citep{wang2023freematch, huang2024dtbs, liu2023hierarchical}, aiming at the mitigation of performance degradation encountered in cases with limited training samples. Commonly, SSL relies on three core assumptions: (1) Smoothness assumption that ensures similar inputs would yield similar outputs and vice versa; (2) Cluster assumption, which suggests that instances of the same class tend to be clustered together in the feature space. Consequently, the classification boundary should traverse sparsely populated regions while avoiding densely populated areas on either side. (3) Manifold assumption, which considers that samples residing within a compact neighborhood in a low-dimensional manifold are likely to share similar labels. Currently, two semi-supervised learning branches, \textit{i.e.}, pseudolabel-based and consistency-based, have been extensively investigated.

\textbf{Pseudolabel-based SSL:} Pseudo-labeling methods adopt a supervised paradigm that simultaneously learns from labeled and unlabeled data. The essence of this branch lies in the reliable generation of pseudo-labels \citep{Lee2013PseudoLabelT}. For instance, Fixmatch \citep{sohn2020fixmatch} employed a fixed threshold to determine the trustworthiness of the samples, ensuring high quality and reliability of the pseudo-labels. Moreover, Ref. \citep{kwon2022semi} utilized an auxiliary error localization network, identifying pixels with potentially erroneous labels. Additionally, Freematch \citep{wang2023freematch} dynamically adjusted the confidence threshold based on the learning states of the involved model.

\textbf{Consistency-based SSL:} According to the assumptions of smoothness and clustering, model predictions are expected to exhibit similarity when specific perturbations are applied, which may involve adjustments to input data, features, or networks. Drawing on this inspiration, Laine et al. introduced a Pi-Model and a temporal ensembling model \citep{pi_modelandtemporal}, aiming to exploit both data-level and model-level consistencies. Subsequently, Tarvainen et al. presented the MT model \citep{tarvainen2017mean}, in which the student network utilizes EMA to update the parameters of teacher network, thereby reducing model-level inconsistencies. To take advantage of unlabeled data, CCT \citep{ouali2020semi} employed the training of multiple auxiliary decoders, each receiving distinct perturbations of the output generated by the shared encoder.


\subsection{Semi-supervised medical image segmentation}
In contrast to natural semantic segmentation, medical images often suffer from limited data availability while requiring higher prediction accuracy. Consequently, there is an urgent need to explore efficient semi-supervised methods to alleviate data requirements and improve accuracy. Through an examination of pseudo-labeling methods, the practical quality can be refined using techniques such as uncertainty knowledge \cite{wang2021semi}, and random propagation \cite{fan2020inf}, among others. Additionally, Lyu et al. \cite{9931157} suggested generating synthetic images aligned with the retained pseudo-labels. In methods employing consistency regularization, Yu et al. \cite{yu2018pu} proposed an uncertainty-aware mean teacher model for left atrium segmentation, while Wang et al. \cite{Wang2020DoubleUncertaintyWM} introduced a double-uncertainty weighted method for semi-supervised applications. Moreover, Huang et al. \cite{huang2022semi} developed a two-stage learning scheme for neuron segmentation, which fully extracts useful information from unlabeled data.
Furthermore, numerous other practices are also available to support semi-supervised medical image segmentation. Bai et al. \citep{Bai2023BidirectionalCF} utilized bidirectional copy-paste to prompt unlabeled data to assimilate comprehensive semantics from labeled data, thus mitigating the experience mismatch problem between labeled and unlabeled data. Additionally, \citep{PseudoLabelGuided, zhao2023rcps, cross-patchContrastive, zhao2022cross} have incorporated contrastive learning into SSMIS, with the aim of learning representations of distinct features and emphasizing differences in feature spaces across various categories.

\vspace{-10pt}

\subsection{Different Structures for SSL}
\vspace{-2pt}
As illustrated in Figure \ref{figure1}, five SSL architectures currently dominate this field, namely MT \citep{tarvainen2017mean}, DS \citep{ke2019dual}, CPS \citep{chen2021semi}, PS-MT \citep{liu2022perturbed}, and UCMT \citep{Shen2023CotrainingWH}. Additionally, our DCF is also swept in generalization.

\textbf{Mean Teacher:} In Figure \ref{figure1} (a), MT model mainly comprises two networks with identical architectures: the teacher network and the student network. While the parameters of the student network are updated through backpropagation, the teacher network undergoes updating via Exponential Moving Average. Nevertheless, this treatment is currently stuck in a performance bottleneck.

\textbf{Dual Student:} DS model involves two students with shared architectures, which utilizes stable samples to impose effective constraints between the two counterparts, thereby mitigating the problem of over-coupling that often encountered within EMA computation. Please refer to Figure \ref{figure1} (b) for a visual depiction.

\textbf{CPS:} As shown in Figure \ref{figure1} (c), the CPS structure integrates both self-training and consistency learning methodologies, wherein one-hot pseudo labels derived from the outcomes of both models serve as supervision signals, mutually guiding and supervising each other's learning processes.

\textbf{PS-MT:} PS-MT in Figure \ref{figure1} (d) employs two teachers. To produce pseudo labels, the prediction outcomes of both are merged using an ensemble approach, bolstering the stability of the pseudo labels. Furthermore, during each training epoch, only one of the teachers undergoes updating, adjusting the model parameters to augment the diversity between two subassemblies.

\textbf{UCMT:} UCMT integrates collaborative mean teacher techniques and uncertainty-guided region mixture to concurrently maintain model inconsistency and high-confidence labels, resulting in promising outcomes. Please see Figure 1(e) for a visual representation.

\textbf{DCF (ours):} Due to EMA computation, MT-based techniques inevitably result in the over-coupling issue, yet the no-teacher alternatives often lack a direct method to enforce consistency constraints. Therefore, we introduce DCF to mitigate these challenges. For an in-depth analysis, please refer to Section \ref{section3}.

\begin{figure}[h]
\centering
\includegraphics[width=0.95\linewidth]{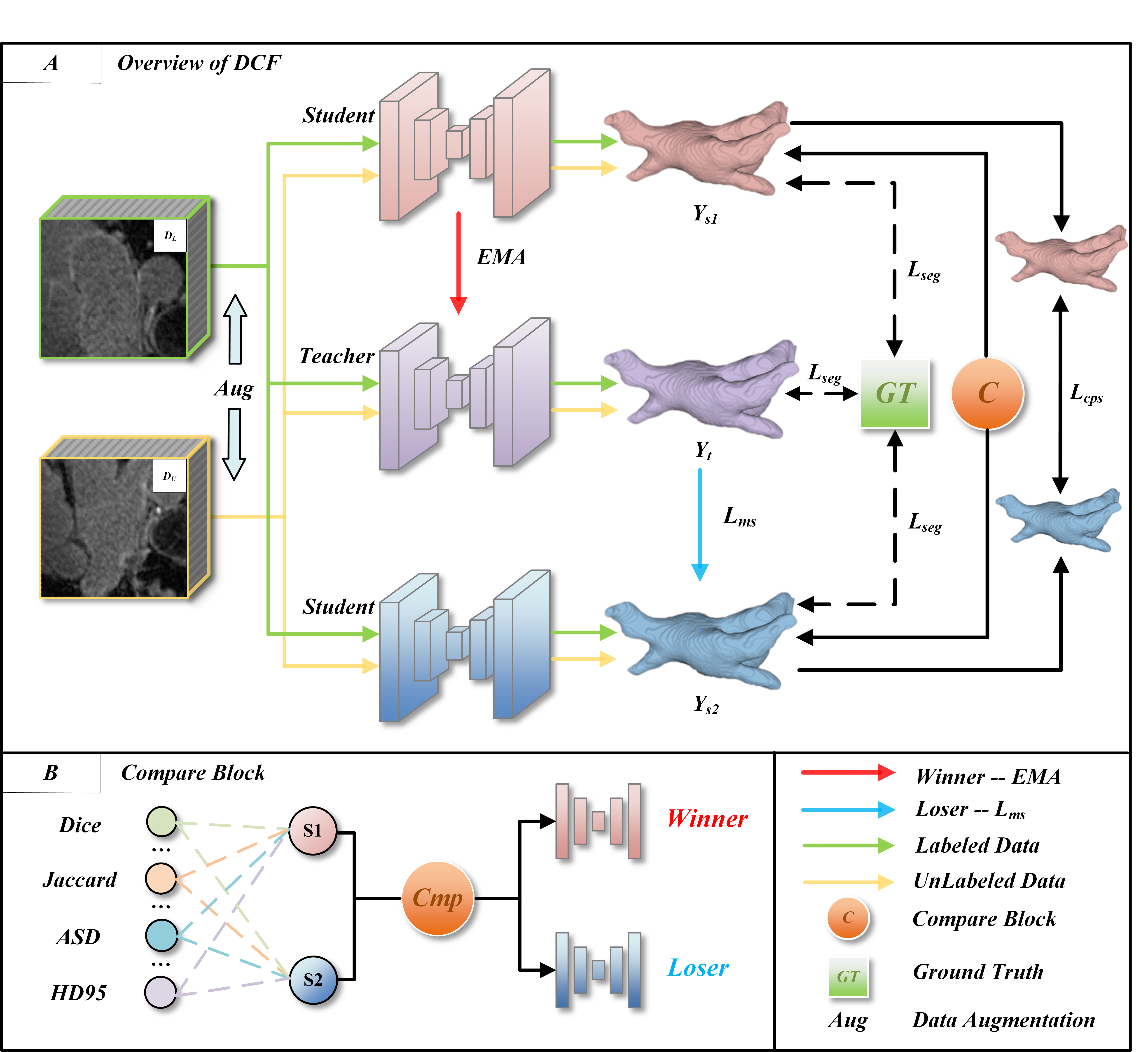}
\caption{Block A introduces our proposed DCF. Upon receiving input data, DCF undergoes two random data augmentations. Then, three separate networks follow: a student network and two teacher networks. Operating on a co-teaching scheme, DCF fosters cross-pseudo supervision between two student results. In Block B, a competitive mechanism employing metrics such as Dice, Cross-entropy, and 95HD is elaborated during training to compare the performance of the two students and determine a winner.}
\label{figure_model}
\end{figure}

\section{Methodology} \label{section3}

\subsection{The overview framework}

During semi-supervised learning, it is assumed that the training dataset contains \( N \) labeled data and \( M \) unlabeled data, where \( M \gg N \). For convenience, we denote the entire training set as \( \mathcal{D} = \{ \mathcal{D}_L, \mathcal{D}_U \} \), with a small portion of labeled data represented as \( \mathcal{D}_L = \{ (x_i^L, y_i^L) \}_{i=1}^N \), and the unlabeled counterpart as \( \mathcal{D}_U = \{ x_{i}^U \}_{i=1}^M \). Here, \( x_i  \) denotes the training image, and \( y_i \) is the label (if available). Yet with a limited number of labeled samples \( x_i^L \), the objective of semi-supervised learning is to achieve promising results with the aid of the extra unlabeled data \( x_i^U \).

As discussed, the MT technique is the dominant framework for most contemporary SSL approaches  \citep{tarvainen2017mean, liu2022perturbed, Bai2023BidirectionalCF, yu2018pu, Fussnet}. However, the Exponential Moving Average (EMA) mechanism leads to excessive coupling between teachers and students, resulting in performance bottlenecks. Subsequently, although the Dual Student framework \citep{ke2019dual, 10120761} has addressed the coupling issue, it requires the development of a stable sample and a training method integrating entra constraints to facilitate the exchange of correct knowledge between the two students and prevent model collapse due to erroneous knowledge exchange.

\begin{figure}[htbp]
	\centering
	\subfigure[Weight Distance]{
            \label{wd}
            \begin{minipage}[b]{0.486\linewidth}
                \includegraphics[width=1\linewidth]{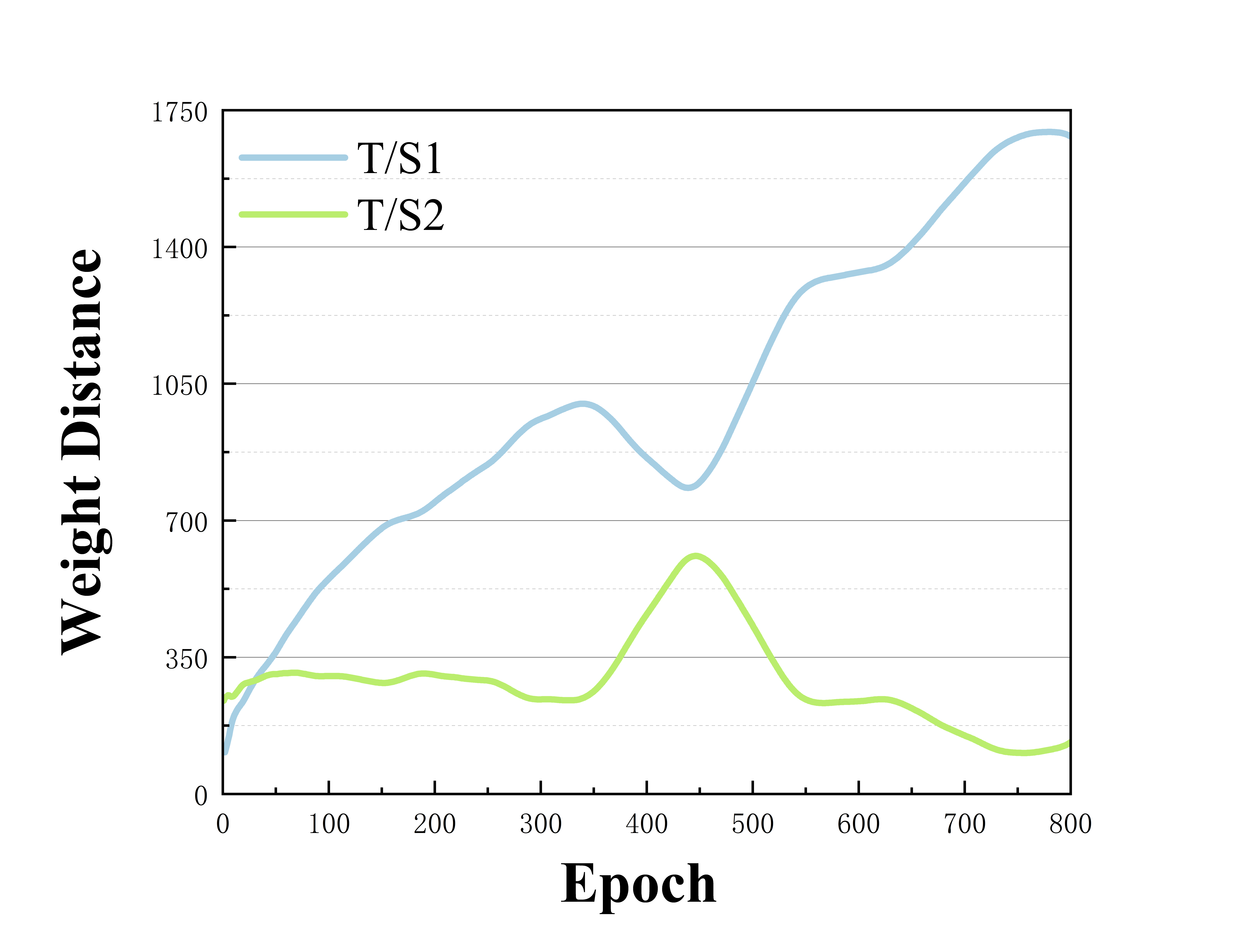}
            \end{minipage}}
    	\centering
	\subfigure[Prediction Distance]{
            \label{pd}
            \begin{minipage}[b]{0.486\linewidth}
                \includegraphics[width=1\linewidth]{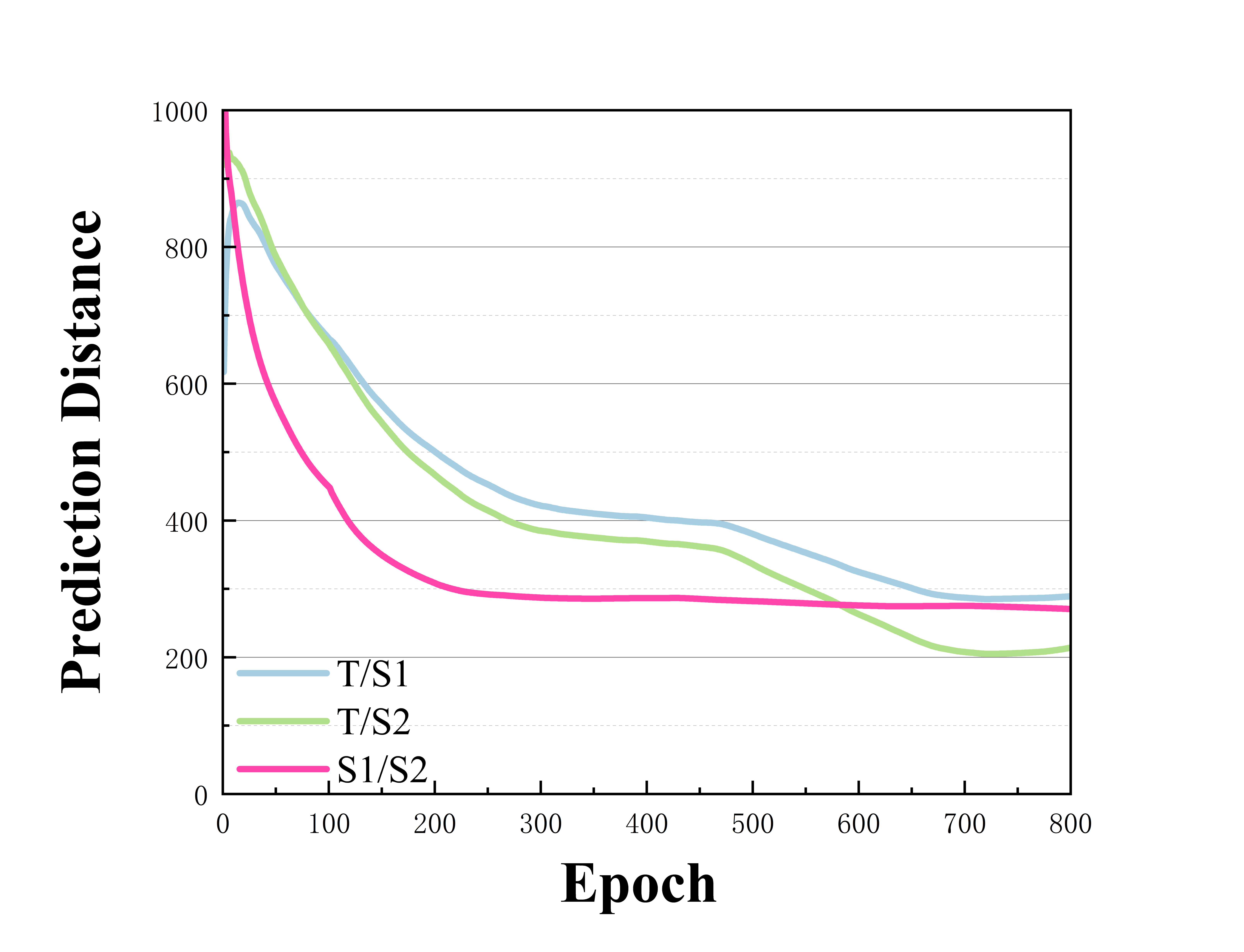}
            \end{minipage}}        
	\caption{(a) denotes the weight distance between teacher and student. (b) represents the prediction distance among three networks. For simplicity, the simple Euclidean distance is used for weight and prediction distances.}
	\label{figure3}
\end{figure}

Motivated by the aforementioned issues, we introduce the Decouple Competitive Framework (DCF), which employs a unique competition mechanism to select students currently with superior performance to effectively mitigate the problem of excessive coupling between a single teacher and a student. Moreover, we further devise a mentoring mechanism, which empowers teachers to grant additional learning privileges to underperforming students, narrowing the cognitive bias between the two students. In addition, we advocate mutual learning and assistance between the two students, facilitated by the teacher. Thus, a straightforward consistency loss between students is sufficient to ensure alignment with learning objectives (detailed elucidation provided in Subsection \ref{3.2}).

The overall framework of the proposed DCF is shown in Figure \ref{figure_model}, which mainly comprises three networks: a teacher network $f (\cdot;\theta_t)$ and two student networks $f (\cdot;\theta_{s1})$ and $f (\cdot;\theta_{s2})$, all initialized randomly. For each training data $X$ that encompasses both labeled and unlabeled samples, we introduce random augmentation $\xi$ and $\xi^{'}$ to generate perturbed instances. Subsequently, these samples independently traverse through the three networks, producing their respective predictions: ${Y}_t = f(X;\theta_t)$, ${Y}_{s1} = f(X+\xi;\theta_{s1})$, and ${Y}_{s2} = f(X+\xi^{'};\theta_{s2})$.

\vspace{-10pt}
\subsection{Decoupled Competitive mechanism} \label{3.2}
\vspace{-2pt}
As analyzed previously, the student model may inadvertently overshadow the teacher model's capacity to assimilate information during the training phase. In the dual student architecture, the learning capabilities of the two students may not be uniform, thus introducing a cognitive bias that can result in suboptimal performance. To address these issues, we propose a straightforward yet effective decoupled competitive mechanism.

The training process of DCF is shown in Algorithm \ref{algorithm}. Referring to previous work \cite{yu2018pu}, for labeled data, cross-entropy loss \( L_{ce} \) and dice loss \( L_{dice} \) are utilized for supervised training:
\begin{equation}
    L_{seg} = L_{ce}(f(x_i^L;\theta), y_i) + L_{dice}(f(x_i^L;\theta), y_i)
\end{equation} 
where $y_i$ is the label of $x_i^L$.

With the supervised loss \( L_{seg} \) calculated using ground truth, a competition function is employed to determine which backbone performs more competitively in the current state. This competition function may encompass various metrics such as Dice coefficient, Cross-entropy (CE), and 95\% Hausdorff distance (95HD), which can be achieved on-the-fly during the training process. Following this line, two advantages can be guaranteed. 1) it obviates the need for modifying the network structure, thus avoiding an increase in parameter count, and 2) as some indicators involve supervised learning processes and must be computed anyway, no additional computational overhead is introduced. Upon determining the winner for the current iteration, we utilize its parameters to update the teacher network $\theta^{'}_t$ at training step $t$ based on EMA, \textit{i.e.}, $\theta^{'}_t = \alpha \theta^{'}_{t-1} + (1-\alpha) \theta_t$, where $\alpha$ is the EMA decay that controls the updating rate. Since the winner changes dynamically, the parameters of teacher network remain uncoupled from any particular students, facilitating the assimilation of more effective information during training. Moreover, from the perspective of the teacher network, the status of both students is the same, hence the decoupling of tight dependency can be accomplished, which further allows for a more exclusive focus on the acquisition of desirable knowledge. 

For unlabeled data, the latent knowledge they harbor merits comprehensive exploration. Hence, we introduce  \( L_{unsup} \) to fully exploit the relationship between labeled and unlabeled data, especially in domains such as medical imaging, where scenes often manifest consistent semantic information across the dataset.

Throughout the overall training process, to effectively promote the efficacy of underperforming students (us), we utilize the teacher network to correctly steer student models toward correct optimization, thereby preventing them from converging in erroneous directions. This can be likened to a tutoring process, for which the mentoring loss is defined as:
\begin{equation}
    L_{ms} = \frac{1}{M} \sum_{j=1}^{M} L_{seg}(f(x_j^U;\theta_{us}), \hat{Y}_j^U)
\end{equation}
where $\hat{Y}_j^U$ is the pseudo label of $f(x_j^U;\theta_{t})$.

As widely acknowledged, maintaining consistency in model prediction results is of paramount importance in semi-supervised learning methods. However, applying conformance constraints directly can cause models to collapse with each other due to the exchange of incorrect knowledge \citep{ke2019dual}. Therefore, Dual Student-based methods often incorporate additional techniques to ensure an accurate exchange of information between models, thereby preventing the collapse issue. Nevertheless, due to the presence of these mechanisms, we can achieve satisfactory results by simply adding a straightforward consistency constraint between the two students, significantly reducing performance overhead. The specific cross-pseudo-supervision loss function is as follows:
\begin{equation}
    L_{cps} = \frac{1}{M} \sum_{j=1}^{M} (L_{seg}(f(x_j^U;\theta_{s1}), \hat{Y}_j^U) + L_{seg}(f(x_j^U;\theta_{s2}), \hat{Y}_j^U))
\end{equation}
where $\hat{Y}j^U$ represents the pseudo label of $f(x_j^U;\theta_{s2})$ if the result compared with him comes from Student1, and vise versa. The loss function for unsupervised data \( L_{unsup} \) can then be formulated as follows:
\begin{equation}
    L_{unsup} = L_{cps} + L_{ms}
\end{equation} 
Note that \( L_{ms} \) is exclusively assigned to currently underperforming students.

With all the sub-loss assembled, the overall loss is given as follows. It's noteworthy that during the early stages of training, the network's uncertainty tends to be relatively high. Therefore, in line with previous practices \citep{zhao2023rcps, luo2021semi}, we introduce a parameter within the \( L_{unsup} \) to stabilize the model training.
\begin{equation}
    L_{total} = L_{seg} + \lambda L_{unsup}
\end{equation} 
where $\lambda$ is the concerned weight for balance control of \( L_{unsup} \).


\begin{algorithm}[tb]
\caption{Training of DCF for SSL}
\label{algorithm}
\textbf{Require}:
\begin{itemize}[leftmargin=1.5em] 
  \item The set of samples: X
  \item The random augmentation: $\xi$, $\xi^{'}$
  \item The teacher network: $f_{t} (\theta_{t})$
  \item The student networks: $f_{s1} (\theta_{s1})$, $f_{s2} (\theta_{s2})$
\end{itemize}
\textbf{Procedure}: 
\begin{algorithmic}[1]
\FOR{each iteration}
\STATE Get $f(X+\xi;\theta_{s1})$, $f(X+\xi^{'};\theta_{s2})$, $f(X;\theta_{t})$
\STATE Calculate supervised Loss on labeled samples
\STATE Calculate cross pseudo supervision loss on unlabeled samples between $f_{s1} (\theta_{s1})$ and $f_{s2} (\theta_{s2})$
\STATE Compare $f_{s1} (\theta_{s1})$, $f_{s2} (\theta_{s2})$ and get the winner $f_{w} (\theta_s)$ and the loser $f_{l} (\theta_s)$
\STATE $f_{t} (\theta_{t})$ assists $f_{l} (\theta_s)$
\STATE $f_{w} (\theta_s)$ updates $f_{t} (\theta_{t})$
\ENDFOR

\end{algorithmic}
\end{algorithm}

\vspace{-10pt}

\subsection{Discussions} \label{3.3}
\vspace{-2pt}
In Figure \ref{wd}, we provide a graphical representation of the weight distance between the teacher network (t) and both student networks (s1 and s2), throughout the training process. To visualize this, we show the curves for more epochs of training. Notably, we observe an inverse relationship between the weight distance of t and s1, and that of t and s2, \textit{i.e.}, as the weight distance between t and s1 decreases, there is a corresponding increase in the weight distance between t and s2, demonstrating a clear antagonistic trend. We attribute this phenomenon to the fact that when s1 performs EMA update on the parameters of t, the weight between them will become similar, resulting in a decrease in the weight distance between s1 and t. This can be interpreted as indicating that the one who updates the teacher with EMA will become closer in weight distance, while the others will show a tendency to move farther away. In this scenario, the weight of the teacher is not overly tethered to a single student, but alternates between two students. As these two students progress in tandem, the teacher can glean effective information from their interaction, thus circumventing the bottleneck of poor performance induced by excessive parameter coupling.

At the same time, we also plot the Prediction Distance between the three networks during the training process, as illustrated in Figure \ref{pd}. It is evident that in the initial stages of training, the prediction distance between the two students steadily diminishes, indicating the influence of the consistency constraint among the students. Consequently, the predicted outcomes of the two students become increasingly similar. Subsequently, the prediction distance between them reaches a threshold and remains relatively constant. Moving forward, the prediction distance between the teacher and the two students exhibits a similar trend, underscoring the model's efficacy and the effective consistency observed among the three networks.

\vspace{-10pt}

\section{Experiments and Results}

\begin{table*}[htbp]
\centering 
\caption{Performance comparison with state-of-the-art methods on LA Dataset. Taking V-Net as the baseline, the green triangle $\textcolor{softgreen}{\blacktriangledown}$ denotes the reduction degree, while upturned red triangle $\textcolor{red}{\blacktriangle}$ represents the rising rate.} 
\label{table1} 
\begin{tabular}{c|cc|cccc}
\hline \multirow{2}{*}{ \textbf{Competing Methods} } & \multicolumn{2}{c|}{\textbf{Volumes used}} & \multicolumn{4}{c}{\textbf{Metrics}} \\  \cline{2-7}
& Labeled & Unlabeled & Dice(\%)$\uparrow$ & Jaccard(\%)$\uparrow$ & 95HD(voxel)$\downarrow$ & ASD(voxel)$\downarrow$  \\
\hline 
V-Net & $8(10\%)$ & $72$  & $78.96$ & $67.82$ & $20.83$ & $5.74$ \\
V-Net & $16(20\%)$ & $64$  & $86.87$ & $77.19$ & $11.93$ & $3.29$ \\
\hline 
UA-MT (MICCAI 2019) & $8(10\%)$ & $72$  & $84.25_{\textcolor{red}{\blacktriangle 6.70\%} }$ & $73.48_{\textcolor{red}{\blacktriangle 8.34 \%} }$ &$13.84_{\textcolor{softgreen}{\blacktriangledown 33.5 \%} }$& $3.36_{\textcolor{softgreen}{\blacktriangledown 41.4 \%} }$ \\
SASSNet (MICCAI 2020) & $8(10\%)$ & $72$  & $87.32_{\textcolor{red}{\blacktriangle 10.6 \%} }$ & $77.72_{\textcolor{red}{\blacktriangle 14.6 \%} }$ &$9.62_{\textcolor{softgreen}{\blacktriangledown 53.8 \%} }$& $2.55_{\textcolor{softgreen}{\blacktriangledown 55.6 \%} }$ \\
DTC (AAAI 2021) & $8(10\%)$ & $72$  & $87.43_{\textcolor{red}{\blacktriangle 10.7 \%} }$ & $78.06_{\textcolor{red}{\blacktriangle 15.1 \%} }$ &$8.37_{\textcolor{softgreen}{\blacktriangledown 59.8 \%} }$& $2.40_{\textcolor{softgreen}{\blacktriangledown  58.2\%} }$ \\
MC-Net+ (MIA 2022) & $8(10\%)$ & $72$  & $88.96_{\textcolor{red}{\blacktriangle 12.7 \%} }$ & $80.25_{\textcolor{red}{\blacktriangle 18.3 \%} }$ &$7.93_{\textcolor{softgreen}{\blacktriangledown 61.9 \%} }$& $1.86_{\textcolor{softgreen}{\blacktriangledown  67.6\%} }$ \\
FUSSNet (MICCAI 2022) & $8(10\%)$ & $72$  & $89.12_{\textcolor{red}{\blacktriangle 12.9 \%} }$ & $80.79_{\textcolor{red}{\blacktriangle 19.1 \%} }$ &$7.13_{\textcolor{softgreen}{\blacktriangledown 65.8 \%} }$& $1.81_{\textcolor{softgreen}{\blacktriangledown  68.5\%} }$ \\
CAML (MICCAI 2023) & $8(10\%)$ & $72$  & $89.44_{\textcolor{red}{\blacktriangle 13.3 \%} }$ & $81.01_{\textcolor{red}{\blacktriangle 19.4 \%} }$ &$10.10_{\textcolor{softgreen}{\blacktriangledown 51.5 \%} }$& $2.09_{\textcolor{softgreen}{\blacktriangledown  63.6\%} }$ \\
UCMT (IJCAI 2023) & $8(10\%)$ & $72$  & $88.13_{\textcolor{red}{\blacktriangle 11.6 \%} }$ & $79.18_{\textcolor{red}{\blacktriangle 16.7 \%} }$ &$9.14_{\textcolor{softgreen}{\blacktriangledown 56.1 \%} }$& $3.06_{\textcolor{softgreen}{\blacktriangledown  46.7\%} }$ \\
VSRC (JBHI 2023) & $8(10\%)$ & $72$  & $88.42_{\textcolor{red}{\blacktriangle 12.0 \%} }$ & $79.57_{\textcolor{red}{\blacktriangle 17.3 \%} }$ &$8.52_{\textcolor{softgreen}{\blacktriangledown 59.1 \%} }$& $2.37_{\textcolor{softgreen}{\blacktriangledown  58.7\%} }$ \\
BCP (CVPR 2023) & $8(10\%)$ & $72$  & $89.62_{\textcolor{red}{\blacktriangle 13.5 \%} }$ & $81.31_{\textcolor{red}{\blacktriangle 19.9 \%} }$ &$6.81_{\textcolor{softgreen}{\blacktriangledown 67.3 \%} }$& $\textbf{1.76}_{\textcolor{softgreen}{\blacktriangledown  69.3\%} }$ \\
DCF (ours) & $8(10\%)$ & $72$  & $\textbf{89.94}_{\textcolor{red}{\blacktriangle 13.9 \%} }$ & $\textbf{81.78}_{\textcolor{red}{\blacktriangle 20.6 \%} }$ &$\textbf{6.38}_{\textcolor{softgreen}{\blacktriangledown 69.4 \%} }$& $1.80_{\textcolor{softgreen}{\blacktriangledown  68.6\%} }$ \\
\hline 
UA-MT (MICCAI 2019) & $16(20\%)$ & $64$  & $88.88_{\textcolor{red}{\blacktriangle 2.31 \%} }$ & $80.21_{\textcolor{red}{\blacktriangle 3.91 \%} }$ &$7.32_{\textcolor{softgreen}{\blacktriangledown 38.6 \%} }$& $2.26_{\textcolor{softgreen}{\blacktriangledown 31.3 \%} }$ \\
SASSNet (MICCAI 2020) & $16(20\%)$ & $64$  & $89.54_{\textcolor{red}{\blacktriangle 3.07 \%} }$ & $81.24_{\textcolor{red}{\blacktriangle 5.25 \%} }$ &$8.24_{\textcolor{softgreen}{\blacktriangledown 30.9 \%} }$& $2.20_{\textcolor{softgreen}{\blacktriangledown 33.1 \%} }$ \\
DTC (AAAI 2021) & $16(20\%)$ & $64$  & $89.42_{\textcolor{red}{\blacktriangle 2.93 \%} }$ & $80.98_{\textcolor{red}{\blacktriangle 4.91 \%} }$ &$7.32_{\textcolor{softgreen}{\blacktriangledown 38.6 \%} }$& $2.10_{\textcolor{softgreen}{\blacktriangledown  36.2\%} }$ \\
MC-Net+ (MIA 2022) & $16(20\%)$ & $64$  & $91.07_{\textcolor{red}{\blacktriangle 4.83 \%} }$ & $83.67_{\textcolor{red}{\blacktriangle 8.39 \%} }$ &$5.84_{\textcolor{softgreen}{\blacktriangledown 51.0 \%} }$& $1.67_{\textcolor{softgreen}{\blacktriangledown  49.2\%} }$ \\
FUSSNet (MICCAI 2022) & $16(20\%)$ & $64$  & $91.13_{\textcolor{red}{\blacktriangle 4.90 \%} }$ & $83.79_{\textcolor{red}{\blacktriangle 8.55 \%} }$ &$\textbf{5.10}_{\textcolor{softgreen}{\blacktriangledown 57.2 \%} }$& $1.56_{\textcolor{softgreen}{\blacktriangledown  52.6\%} }$ \\
CAML (MICCAI 2023) & $16(20\%)$ & $64$  & $90.71_{\textcolor{red}{\blacktriangle 4.42 \%} }$ & $83.08_{\textcolor{red}{\blacktriangle 7.63 \%} }$ &$6.08_{\textcolor{softgreen}{\blacktriangledown 49.0 \%} }$& $1.59_{\textcolor{softgreen}{\blacktriangledown  51.7\%} }$ \\
UCMT (IJCAI 2023) & $16(20\%)$ & $64$  & $90.41_{\textcolor{red}{\blacktriangle 4.07 \%} }$ & $82.54_{\textcolor{red}{\blacktriangle 6.93 \%} }$ &$6.31_{\textcolor{softgreen}{\blacktriangledown 47.1 \%} }$& $1.70_{\textcolor{softgreen}{\blacktriangledown  48.3\%} }$ \\
VSRC (JBHI 2023) & $16(20\%)$ & $64$  & $90.59_{\textcolor{red}{\blacktriangle 4.28 \%} }$ & $82.60_{\textcolor{red}{\blacktriangle 7.01 \%} }$ &$5.60_{\textcolor{softgreen}{\blacktriangledown 53.1 \%} }$& $1.72 _{\textcolor{softgreen}{\blacktriangledown  47.7\%} }$ \\
BCP (CVPR 2023) & $16(20\%)$ & $64$  & $90.74_{\textcolor{red}{\blacktriangle 4.45 \%} }$ & $83.17_{\textcolor{red}{\blacktriangle 7.75 \%} }$ &$6.40_{\textcolor{softgreen}{\blacktriangledown 46.3 \%} }$& $1.65_{\textcolor{softgreen}{\blacktriangledown  49.8\%} }$ \\
DCF (ours) & $16(20\%)$ & $64$  & $\textbf{91.44}_{\textcolor{red}{\blacktriangle 5.26 \%} }$ & $\textbf{84.28}_{\textcolor{red}{\blacktriangle 9.19 \%} }$ &$5.24_{\textcolor{softgreen}{\blacktriangledown 56.1 \%} }$& $\textbf{1.55}_{\textcolor{softgreen}{\blacktriangledown  52.9\%} }$ \\
\hline
\end{tabular}
\end{table*}

\subsection{Datasets and Metrics}

\textbf{ISIC Dataset.} ISIC \citep{codella2018skin} was released by the International Skin Imaging Collaboration (ISIC), which comprises 2594 dermoscopic 2D images along with the corresponding annotations. Following \citep{ijcai2022p201, Shen2023CotrainingWH}, we use 1815 images for training and 779 images for validation. In the training set, 5\% (91) and 10\% (181) of the images are labeled for different semi-supervised experimental settings.

\textbf{Left Atrial (LA) Dataset.} LA \cite{xiong2020global} is a benchmark dataset from the 2018 Atrial Segmentation Challenge, consisting of 100 3D gadolinium-enhanced MR imaging volumes. Each volume has an isotropic resolution of \(0.625 \times 0.625 \times 0.625 mm^3\), whose ground truth labels are all given. According to previous work \citep{yu2018pu}, we utilize 80 scans for training purposes and reserve 20 scans for evaluation. In the training set, 10\% (8), and 20\% (16) of the images are labeled for different semi-supervised experimental settings.

\begin{figure}[h]
\centering
\includegraphics[width=0.95\linewidth]{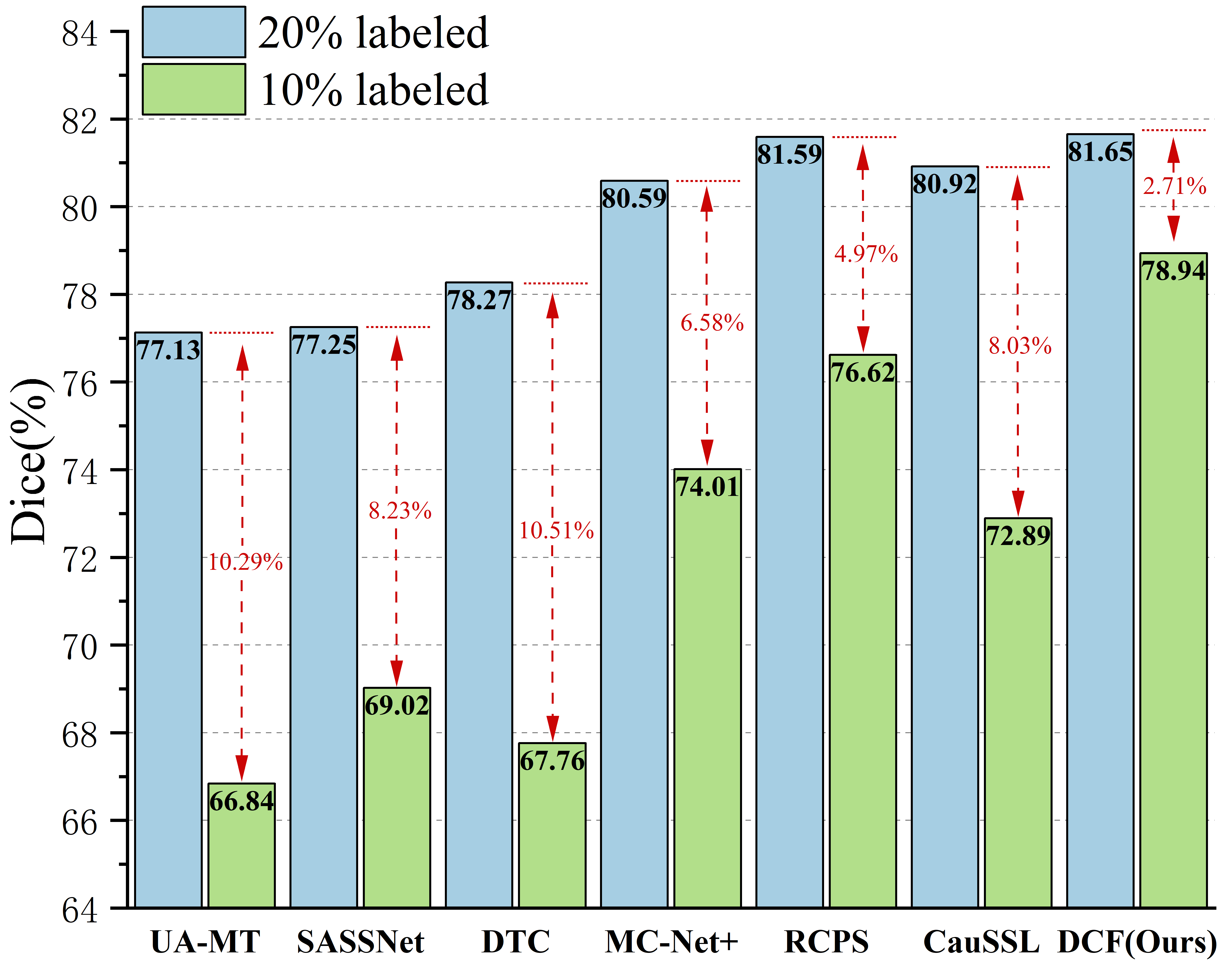}
\caption{Dice scores for 10\% and 20\% labeled data across various models on the Pancreas-CT dataset. Our method exhibits a significantly smaller performance gap between these two cases.}
\label{figure_zzt}
\end{figure}

\textbf{Pancreas-CT Dataset.} Pancreas-CT is also a well-known dataset \cite{Roth2015DeepOrganMD}, which is publicly accessible from the National Institutes of Health Clinical Center. For ease of research and analysis, the scans were preprocessed, involving adjustments of Hounsfield Units (HU) to ranges of \([-125, 275]\) or \([-120, 240]\), as per the specific study requirements, and resampled to an isotropic resolution of \(1.0\ mm \times 1.0\ mm \times 1.0\ mm\). Consistent with previous protocols \citep{wu2022mutual, Miao2023CauSSLCS}, the dataset is divided into 62 training samples and 20 samples for performance evaluation.

\begin{table*}[htbp] %
\centering %
\caption{Performance comparison with state-of-the-art methods on Pancreas-CT Dataset. Taking V-Net as the baseline, the green triangle $\textcolor{softgreen}{\blacktriangledown}$ denotes the reduction degree, while upturned red triangle $\textcolor{red}{\blacktriangle}$ represents the rising rate.} 
\label{table2}
\begin{tabular}{c|cc|cccc}
\hline 
\multirow{2}{*}{ \textbf{Competing Methods} } & \multicolumn{2}{c|}{\textbf{Volumes used}} & \multicolumn{4}{c}{\textbf{Metrics}} \\  
\cline{2-7}
& Labeled & Unlabeled & Dice(\%)$\uparrow$ & Jaccard(\%)$\uparrow$ & 95HD(voxel)$\downarrow$ & ASD(voxel)$\downarrow$  \\
\hline 
V-Net & $6(10\%)$ & $56$  & $55.10$ & $41.02$ &$33.72$ & $12.79$ \\
V-Net & $12(20\%)$ & $50$  & $72.24$ & $58.22$ & $19.39$ & $5.39$ \\
\hline 
UA-MT (MICCAI 2019) & $6(10\%)$ & $56$  & $66.84_{\textcolor{red}{\blacktriangle 21.3 \%} }$ & $51.73_{\textcolor{red}{\blacktriangle 26.1 \%} }$ &$21.32_{\textcolor{softgreen}{\blacktriangledown 36.8 \%} }$& $6.12_{\textcolor{softgreen}{\blacktriangledown 52.2 \%} }$ \\
SASSNet (MICCAI 2020) & $6(10\%)$ & $56$  & $69.02_{\textcolor{red}{\blacktriangle 25.3 \%} }$ & $53.21_{\textcolor{red}{\blacktriangle 29.7 \%} }$ &$18.77_{\textcolor{softgreen}{\blacktriangledown 44.3 \%} }$& $3.09_{\textcolor{softgreen}{\blacktriangledown 75.8 \%} }$ \\
DTC (AAAI 2021) & $6(10\%)$ & $56$  & $67.76_{\textcolor{red}{\blacktriangle 23.0 \%} }$ & $52.14_{\textcolor{red}{\blacktriangle 27.1 \%} }$ &$15.98_{\textcolor{softgreen}{\blacktriangledown 52.6 \%} }$& $4.21_{\textcolor{softgreen}{\blacktriangledown  67.1\%} }$ \\
MC-Net+ (MIA 2022) & $6(10\%)$ & $56$  & $74.01_{\textcolor{red}{\blacktriangle 34.3 \%} }$ & $60.02_{\textcolor{red}{\blacktriangle 46.3 \%} }$ &$12.59_{\textcolor{softgreen}{\blacktriangledown 62.7 \%} }$& $3.34_{\textcolor{softgreen}{\blacktriangledown  73.9\%} }$ \\
RCPS (JBHI 2023) & $6(10\%)$ & $56$  & $76.62_{\textcolor{red}{\blacktriangle 39.1 \%} }$ & $62.96_{\textcolor{red}{\blacktriangle 53.5 \%} }$ &$16.32_{\textcolor{softgreen}{\blacktriangledown 51.6 \%} }$& $3.01_{\textcolor{softgreen}{\blacktriangledown  76.5\%} }$ \\
CauSSL (ICCV 2023) & $6(10\%)$ & $56$  & $72.89_{\textcolor{red}{\blacktriangle 32.3 \%} }$ & $58.06_{\textcolor{red}{\blacktriangle 41.5 \%} }$ &$14.19_{\textcolor{softgreen}{\blacktriangledown 57.9 \%} }$& $4.37_{\textcolor{softgreen}{\blacktriangledown  65.8\%} }$ \\
DCF (ours) & $6(10\%)$ & $56$  & $\textbf{78.94}_{\textcolor{red}{\blacktriangle 43.3 \%} }$ & $\textbf{66.05}_{\textcolor{red}{\blacktriangle 61.0 \%} }$ &$\textbf{11.69}_{\textcolor{softgreen}{\blacktriangledown 65.3 \%} }$& $\textbf{1.38}_{\textcolor{softgreen}{\blacktriangledown 89.2\%} }$ \\
\hline 
UA-MT (MICCAI 2019) & $12(20\%)$ & $50$  & $77.13_{\textcolor{red}{\blacktriangle 6.77 \%} }$ & $63.28_{\textcolor{red}{\blacktriangle 8.70 \%} }$ &$10.52_{\textcolor{softgreen}{\blacktriangledown 45.7 \%} }$& $2.39_{\textcolor{softgreen}{\blacktriangledown 55.6 \%} }$ \\
SASSNet (MICCAI 2020) & $12(20\%)$ & $50$  & $77.25_{\textcolor{red}{\blacktriangle 6.94 \%} }$ & $63.59_{\textcolor{red}{\blacktriangle 9.22 \%} }$ &$11.98_{\textcolor{softgreen}{\blacktriangledown 38.2 \%} }$& $3.12_{\textcolor{softgreen}{\blacktriangledown 42.1 \%} }$ \\
DTC (AAAI 2021) & $12(20\%)$ & $50$  & $78.27_{\textcolor{red}{\blacktriangle 8.35 \%} }$ & $64.75_{\textcolor{red}{\blacktriangle 11.2 \%} }$ &$8.36_{\textcolor{softgreen}{\blacktriangledown 56.9 \%} }$& $2.25_{\textcolor{softgreen}{\blacktriangledown  58.2\%} }$ \\
MC-Net+ (MIA 2022) & $12(20\%)$ & $50$  & $80.59_{\textcolor{red}{\blacktriangle 11.6 \%} }$ & $68.08_{\textcolor{red}{\blacktriangle 16.9 \%} }$ &$\textbf{6.47}_{\textcolor{softgreen}{\blacktriangledown 66.3 \%} }$& $1.74_{\textcolor{softgreen}{\blacktriangledown  67.7\%} }$ \\
RCPS (JBHI 2023) & $12(20\%)$ & $50$  & $81.59_{\textcolor{red}{\blacktriangle 12.9 \%} }$ & $69.04_{\textcolor{red}{\blacktriangle 18.6 \%} }$ &$7.50_{\textcolor{softgreen}{\blacktriangledown 61.3 \%} }$& $2.03_{\textcolor{softgreen}{\blacktriangledown  62.3\%} }$ \\
CauSSL (ICCV 2023) & $12(20\%)$ & $50$  & $80.92_{\textcolor{red}{\blacktriangle 12.0 \%} }$ & $68.26_{\textcolor{red}{\blacktriangle 17.2 \%} }$ &$8.11_{\textcolor{softgreen}{\blacktriangledown 58.2 \%} }$& $1.53_{\textcolor{softgreen}{\blacktriangledown  71.6\%} }$ \\
DCF (ours) & $12(20\%)$ & $50$  & $\textbf{81.65}_{\textcolor{red}{\blacktriangle 13.0 \%} }$ & $\textbf{69.48}_{\textcolor{red}{\blacktriangle 19.3 \%} }$ &$6.77_{\textcolor{softgreen}{\blacktriangledown 65.1 \%} }$& $\textbf{1.21}_{\textcolor{softgreen}{\blacktriangledown  77.6\%} }$ \\
\hline
\end{tabular}
\end{table*}

\textbf{Evaluation Metrics:} For 3D datasets, four typical metrics with different criteria are employed, including Dice Similarity Coefficient (Dice), Jaccard Similarity Coefficient (Jac), 95\% Hausdorff Distance (95HD), and Average Surface Distance (ASD). Among them, Dice and Jaccard are regional sensitivity metrics assessing the overlap between predictions and ground truth. Both 95HD and ASD are edge-sensitive metrics. The former determines the maximum surface-to-surface distance at the 95th percentile between predicted and actual regions, while the latter computes the average distance between concerned points on both surfaces. As for 2D datasets, the primary evaluation metric for segmentation performance is the widely used Dice coefficient.

\vspace{-10pt}

\subsection{Implementation Details}
\vspace{-2pt}
DCF is implemented with PyTorch and executed on an NVIDIA 3090 GPU. In the following, we provide distinct processing methods employed for various datasets. 

\textbf{Left Atrial Dataset:} For the LA dataset, we utilize Vnet as the baseline and trained the network for 500 epochs. The batch size is set to four, comprising two labeled and two unlabeled images. During the training phase, random volume cropping is conducted, resulting in input dimensions of \(112 \times 112 \times 80\) for model updates. During the inference phase, segmentation results are generated using a sliding window with the same dimensions and a stride of \(18 \times 18 \times 4.\). AdamW is employed as the optimizer, with a fixed learning rate of 1e-4.

\textbf{Pancreas-CT Dataset:} Throughout the training process, all volumes undergo random cropping to attain dimensions of \(96 \times 96 \times 96\). While during inference, a stride of \(16 \times 16 \times 16\) is implemented. We trained the network for 600 epochs for the PA dataset. Other configurations mirror those of the LA dataset.

\textbf{ISIC Dataset:} DeepLabv3+ augmented with ResNet50 serves as the baseline architecture for the ISIC dataset. The batch size is set to 8, including 4 labeled samples and 4 unlabeled samples. All images are resized to 256×256 during inference, with outputs reverted to their original dimensions for evaluation. Again, AdamW serves as the optimizer, with a fixed learning rate set at 1e-4. We train the network for 30 epochs for the ISIC dataset.

\vspace{-10pt}

\subsection{Results on Left Atrial Dataset}
\vspace{-2pt}
The evaluation results on LA are summarized in Table \ref{table1}, where we compare our proposed DCF with several other SSMIS methods, including UA-MT \citep{yu2018pu}, SASSNet \citep{li2020shape}, DTC \citep{luo2021semi}, MC-Net+ \citep{wu2022mutual}, FUSSNet \citep{Fussnet}, CAML \citep{gao2023correlation}, UCMT \citep{Shen2023CotrainingWH}, VSRC \citep{10120761}, and BCP \citep{Bai2023BidirectionalCF}. Additionally, the classic V-net is used as a fully supervised benchmark model, presenting its performance for reference purposes. To ensure a fair comparison, we implement these models using their official codes and maintain consistency with their respective parameter settings. Furthermore, to comprehensively assess the performance across varying degrees of supervision, we employ 8 (10\% supervision), and 16 (20\% supervision) samples from the training dataset as labeled data, while treating the remainings as unlabeled data.

\begin{figure*}[htbp]
	\centering
	\subfigure[UAMT]{
            \begin{minipage}[b]{0.103\linewidth}
                \includegraphics[width=1\linewidth]{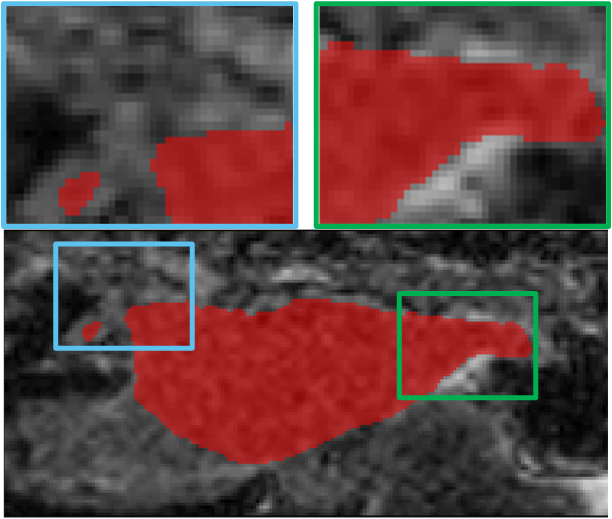}\vspace{5pt}
                \includegraphics[width=1\linewidth]{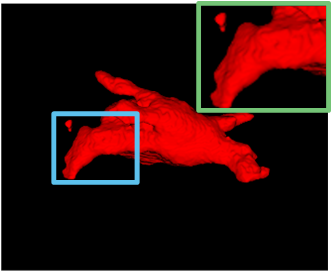} 
            \end{minipage}}
    	\centering
	\subfigure[SASSNet]{
            \begin{minipage}[b]{0.103\linewidth}
                \includegraphics[width=1\linewidth]{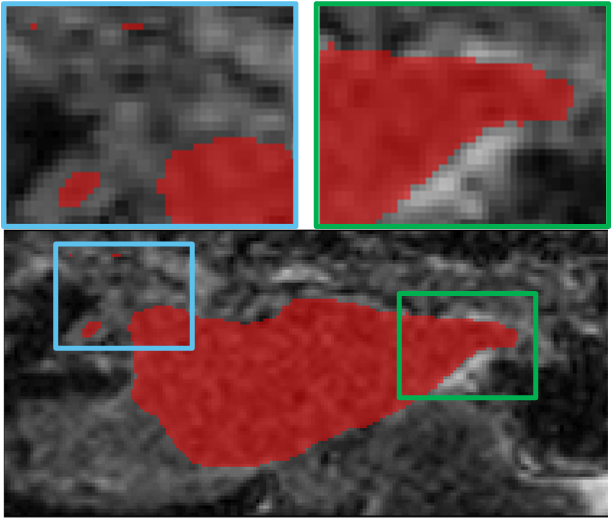}\vspace{5pt}
                \includegraphics[width=1\linewidth]{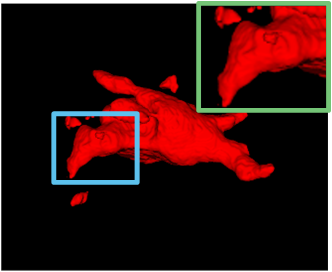} 
            \end{minipage}}
        	\centering
	\subfigure[DTC]{
            \begin{minipage}[b]{0.103\linewidth}
                \includegraphics[width=1\linewidth]{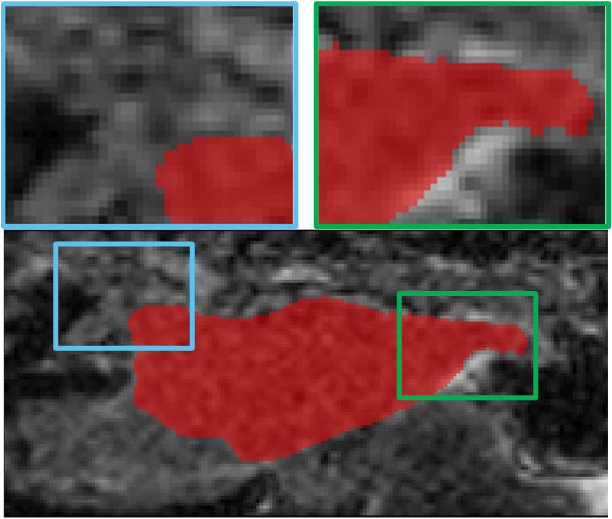}\vspace{5pt}
                \includegraphics[width=1\linewidth]{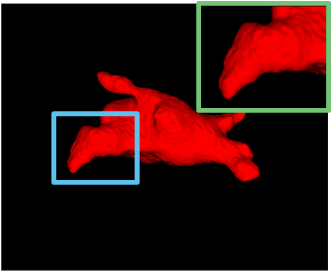} 
            \end{minipage}}
        	\centering
	\subfigure[MC-Net+]{
            \begin{minipage}[b]{0.103\linewidth}
                \includegraphics[width=1\linewidth]{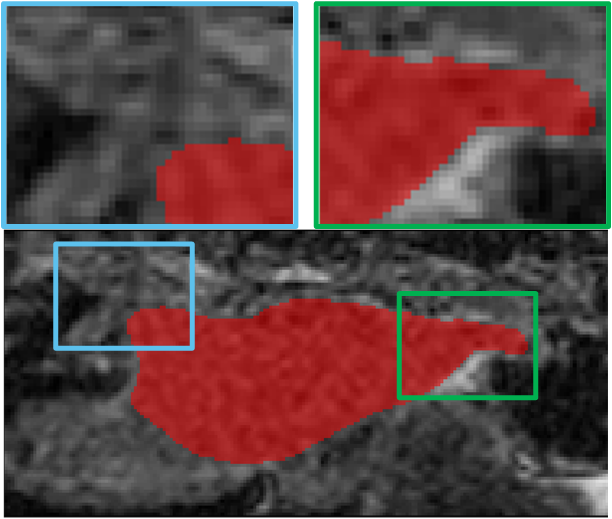}\vspace{5pt}
                \includegraphics[width=1\linewidth]{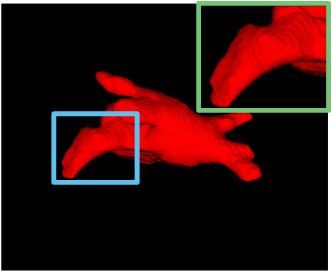} 
            \end{minipage}}
        	\centering
	\subfigure[FUSSNet]{
            \begin{minipage}[b]{0.103\linewidth}
                \includegraphics[width=1\linewidth]{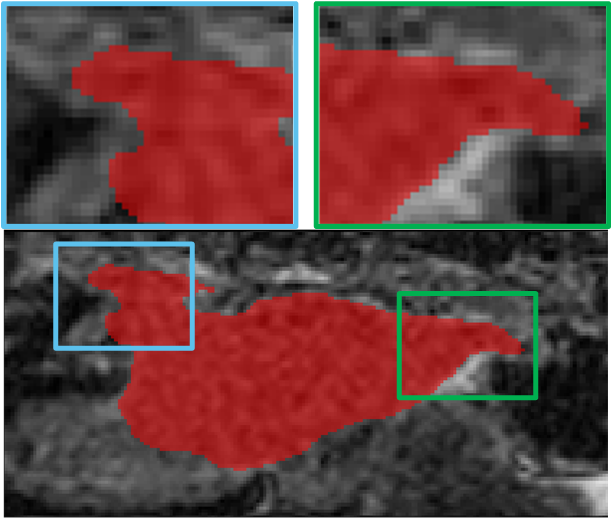}\vspace{5pt}
                \includegraphics[width=1\linewidth]{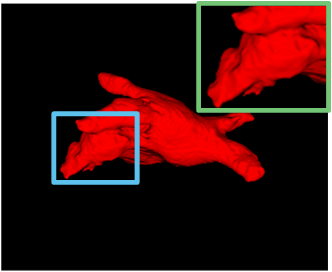} 
            \end{minipage}}
        	\centering
	\subfigure[CAML]{
            \begin{minipage}[b]{0.103\linewidth}
                \includegraphics[width=1\linewidth]{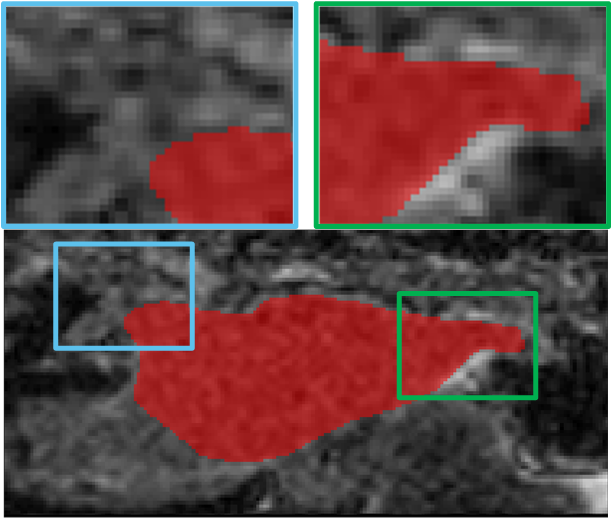}\vspace{5pt}
                \includegraphics[width=1\linewidth]{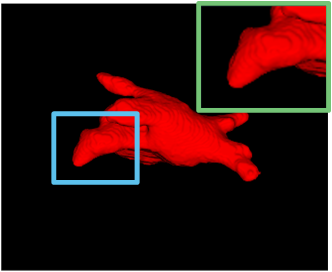} 
            \end{minipage}}
        	\centering
	\subfigure[VSRC]{
            \begin{minipage}[b]{0.103\linewidth}
                \includegraphics[width=1\linewidth]{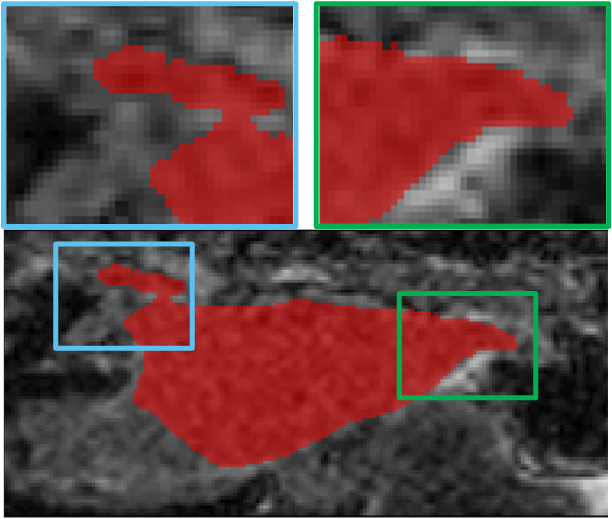}\vspace{5pt}
                \includegraphics[width=1\linewidth]{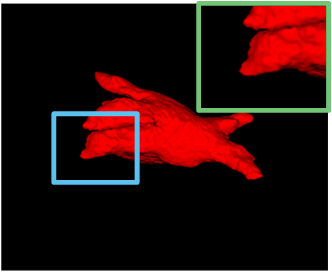} 
            \end{minipage}}
        	\centering
	\subfigure[Ours]{
            \begin{minipage}[b]{0.103\linewidth}
                \includegraphics[width=1\linewidth]{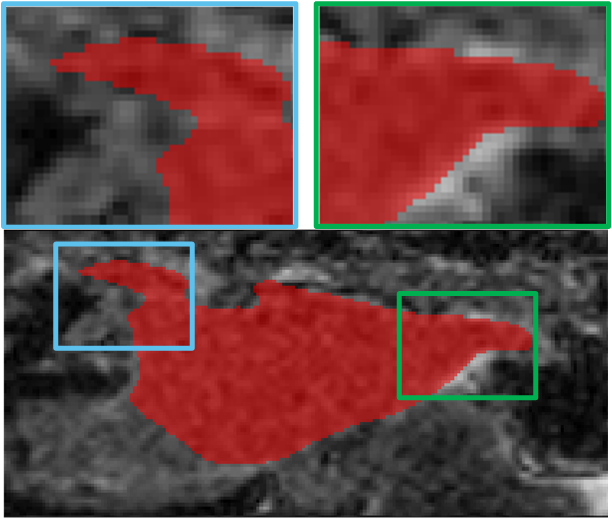}\vspace{5pt}
                \includegraphics[width=1\linewidth]{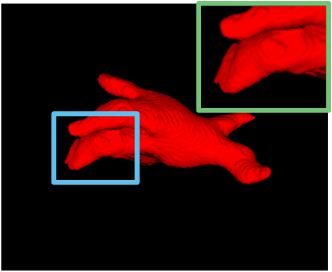} 
            \end{minipage}}
        \centering
	\subfigure[Ground Truth]{
            \begin{minipage}[b]{0.103\linewidth}
                \includegraphics[width=1\linewidth]{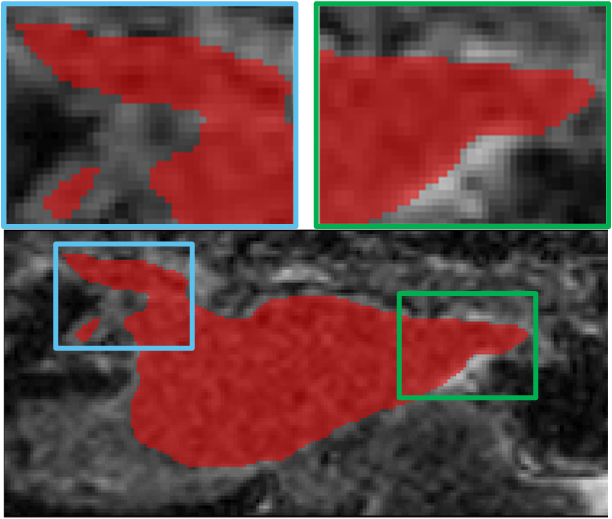}\vspace{5pt}
                \includegraphics[width=1\linewidth]{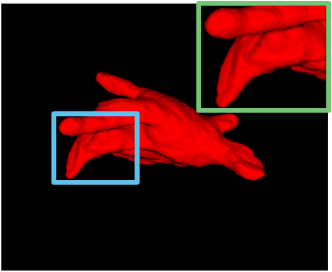} 
            \end{minipage}}
            
	\caption{Comparison of visualization results on LA Dataset. The first row shows the results in 2D form, while the second row provides the visualizations in 3D form, where certain details have been enlarged for better clarity.}
	\label{figure2}
\end{figure*}

In the table presented, our proposal exhibits a notable advancement in Dice score, elevating from 78.96\% to 89.94\% when utilizing only 10\% labeled data, showcasing a distinct advantage over alternative methodologies. With a subsequent increase in labeled data to 20\%, DCF further boosts the performance to 91.44\%, marking a notable gap of 4.57\% compared to the baseline. In particular, DCF consistently outperforms the other competing approaches in both supervision settings, underscoring its superiority. Furthermore, for a more visually comprehensible depiction of segmentation outcomes, Figure \ref{figure2} illustrates the results on the LA dataset. At first glance, it can be easily observed that our approach yields clearer segmentation boundaries and finer-grained details, aligning closely with the Ground Truth.

\vspace{-10pt}

\subsection{Result on Pancreas-CT Dataset}
\vspace{-2pt}
Table \ref{table2} shows the results specific to Pancreas-CT. Note that the volumes in this dataset provide a more complex backdrop compared to LA MRIs, rendering pancreas segmentation a more challenging task. To facilitate an intuitive comparison, we again employ several state-of-the-art competitors, namely UA-MT \citep{yu2018pu}, SASSNet \citep{li2020shape}, DTC \citep{luo2021semi}, MC-Net+ \citep{wu2022mutual}, RCPS \citep{zhao2023rcps}, and CauSSL \citep{Miao2023CauSSLCS}. The performance metrics reported in their respective papers are directly adopted. Similarly, we employ Vnet with varying proportions of labeled data (10\%, and 20\%) for comparative analysis. Similarly, we select 6 samples for 10\% supervision and 12 samples for 20\% supervision from the training dataset, and consider the remainder as unlabeled data. 

\begin{table}[htbp]
\centering 
\caption{Performance comparison with state-of-the-art methods on ISIC Dataset. The best results are in Bold font.} 
\label{table3} 
\begin{tabular}{c|cc|c}
\hline  
\multirow{2}{*}{\textbf{Methods}} & \multicolumn{2}{c|}{\textbf{Volumes used}} & \multicolumn{1}{c}{\textbf{Metrics}} \\  
\cline{2-4}
& Labeled & Unlabeled & Dice(\%)$\uparrow$  \\
\hline 
MT (NIPS 2017) & $90(5\%)$  &$1725$ & $86.43$ \\
UA-MT (MICCAI 2019) & $90(5\%)$  &$1725$ & $87.02$ \\
CCT (CVPR 2020) & $90(5\%)$  &$1725$ & $84.48$ \\
CPS (CVPR 2021) & $90(5\%)$  &$1725$ & $86.79$ \\
UCMT (IJCAI 2023) & $90(5\%)$  &$1725$ & $88.22$ \\
DCF (ours) & $90(5\%)$  &$1725$ & $\textbf{88.88}$ \\
\hline 
MT (NIPS 2017) & $181(10\%)$  &$1634$ & $86.97$ \\
UA-MT (MICCAI 2019) & $181(10\%)$  &$1634$ & $87.48$ \\
CCT (CVPR 2020) & $181(10\%)$  &$1634$ & $85.72$ \\
CPS (CVPR 2021) & $181(10\%)$  &$1634$ & $87.92$ \\
UCMT (IJCAI 2023) & $181(10\%)$  &$1634$ & $88.46$ \\
DCF (ours) & $181(10\%)$  &$1634$ & $\textbf{89.23}$ \\
\hline
\end{tabular}
\end{table}

\begin{figure*}[htbp]
	\centering
	\subfigure[UAMT]{
            \begin{minipage}[b]{0.116\linewidth}
                \includegraphics[width=1\linewidth]{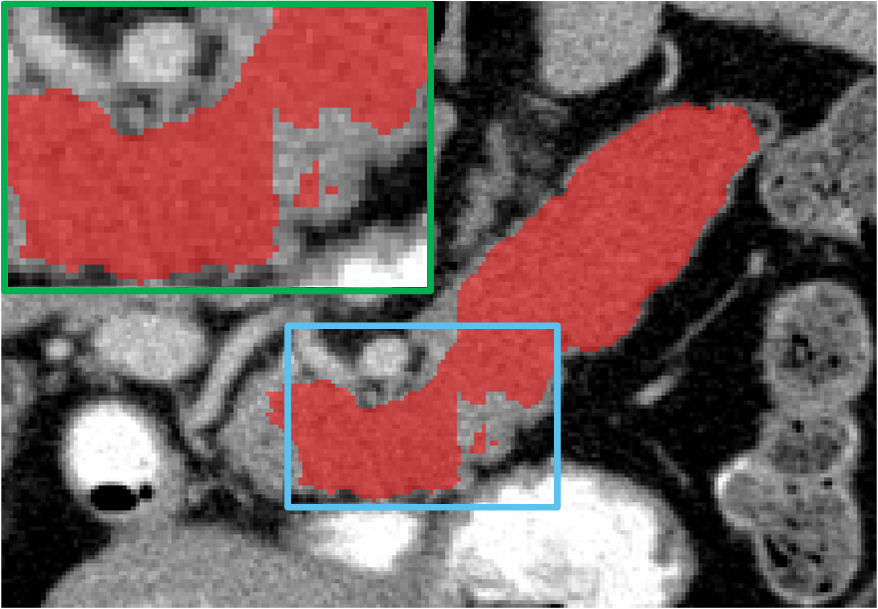}\vspace{5pt}
                \includegraphics[width=1\linewidth]{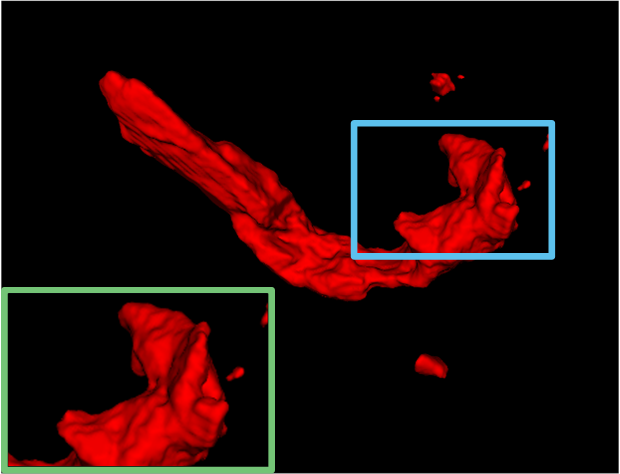} 
            \end{minipage}}
    	\centering
	\subfigure[SASSNet]{
            \begin{minipage}[b]{0.116\linewidth}
                \includegraphics[width=1\linewidth]{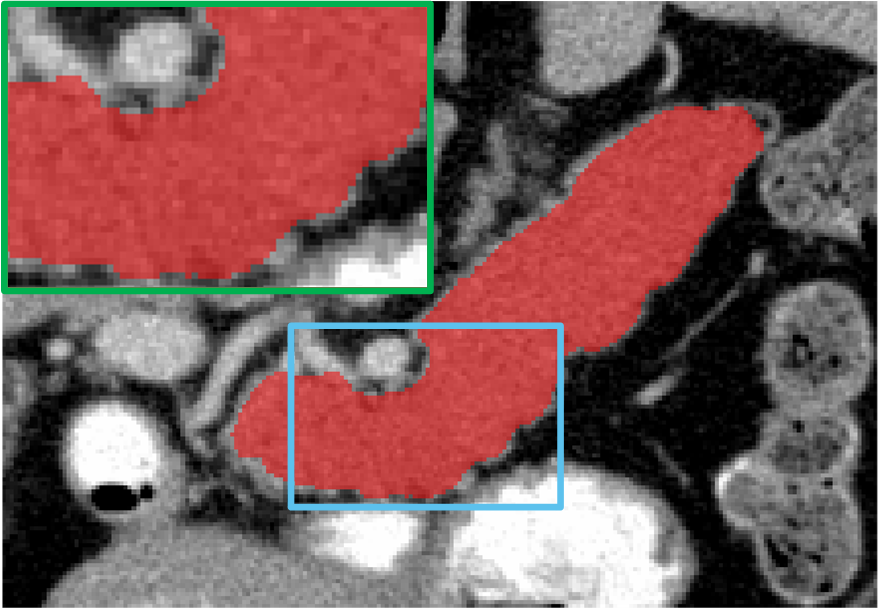}\vspace{5pt}
                \includegraphics[width=1\linewidth]{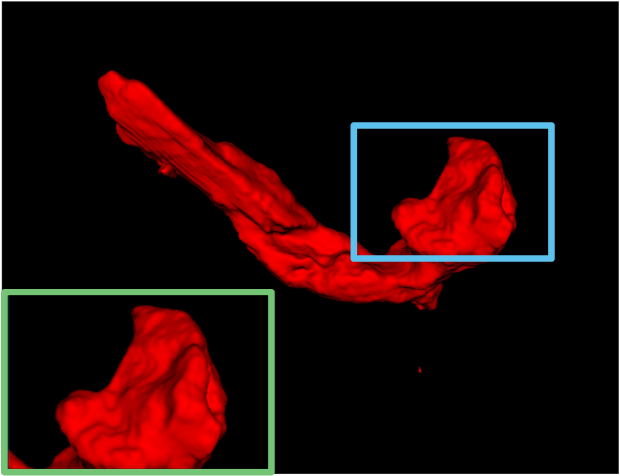} 
            \end{minipage}}
        	\centering
	\subfigure[DTC]{
            \begin{minipage}[b]{0.116\linewidth}
                \includegraphics[width=1\linewidth]{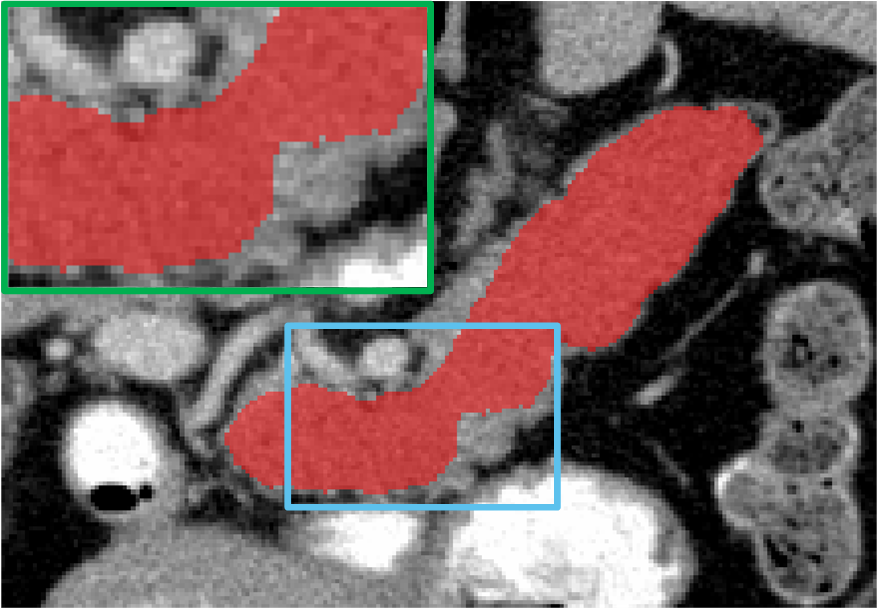}\vspace{5pt}
                \includegraphics[width=1\linewidth]{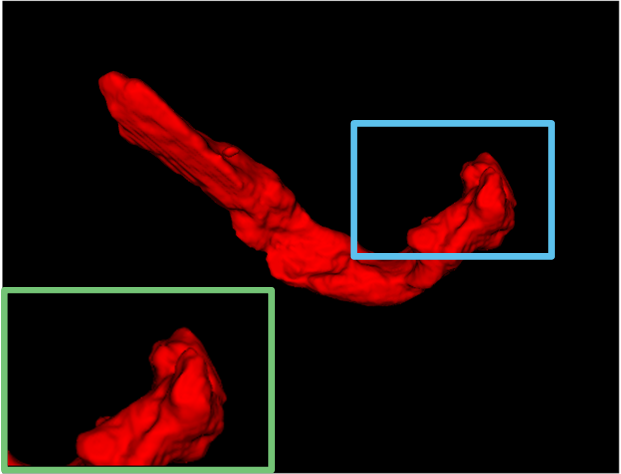} 
            \end{minipage}}
        	\centering
	\subfigure[MC-Net+]{
            \begin{minipage}[b]{0.116\linewidth}
                \includegraphics[width=1\linewidth]{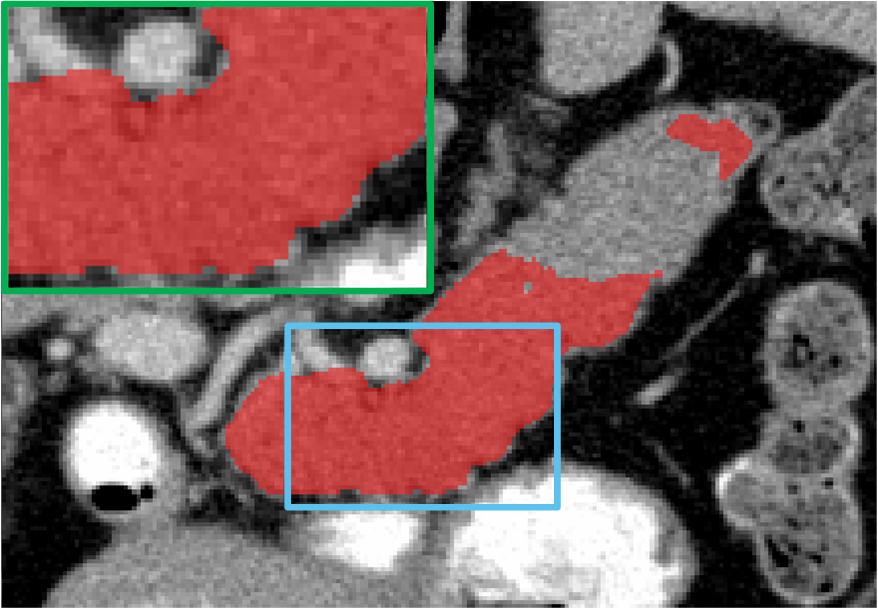}\vspace{5pt}
                \includegraphics[width=1\linewidth]{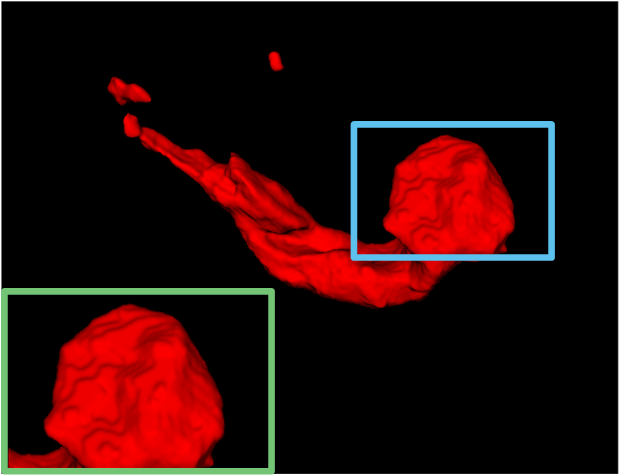} 
            \end{minipage}}
        	\centering
	\subfigure[RCPS]{
            \begin{minipage}[b]{0.116\linewidth}
                \includegraphics[width=1\linewidth]{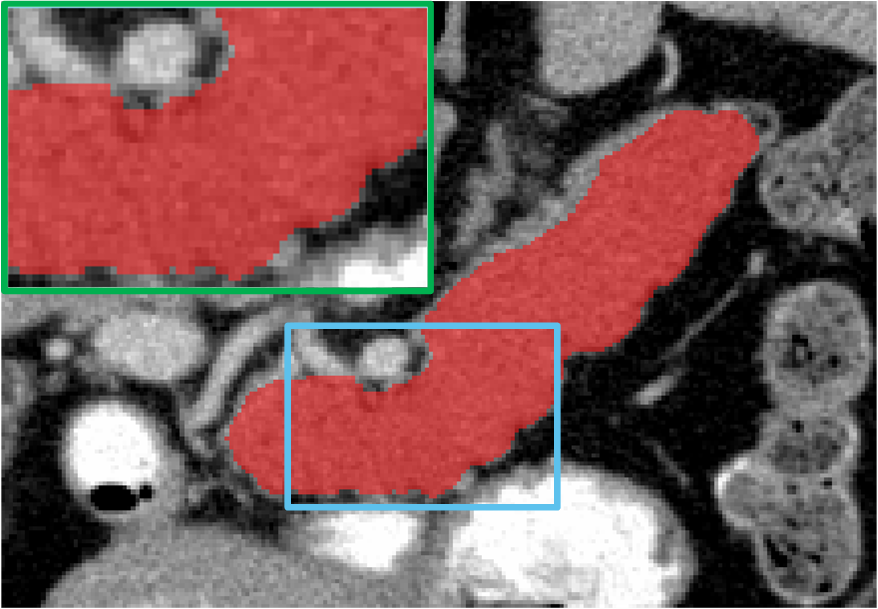}\vspace{5pt}
                \includegraphics[width=1\linewidth]{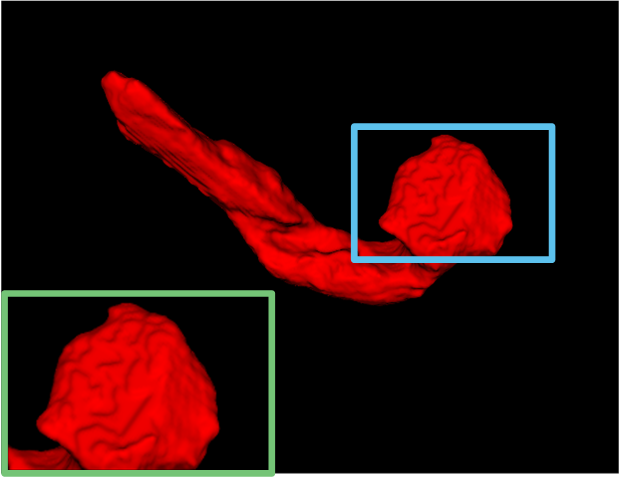} 
            \end{minipage}}
        	\centering
         \subfigure[CauSSL]{
            \begin{minipage}[b]{0.116\linewidth}
                \includegraphics[width=1\linewidth]{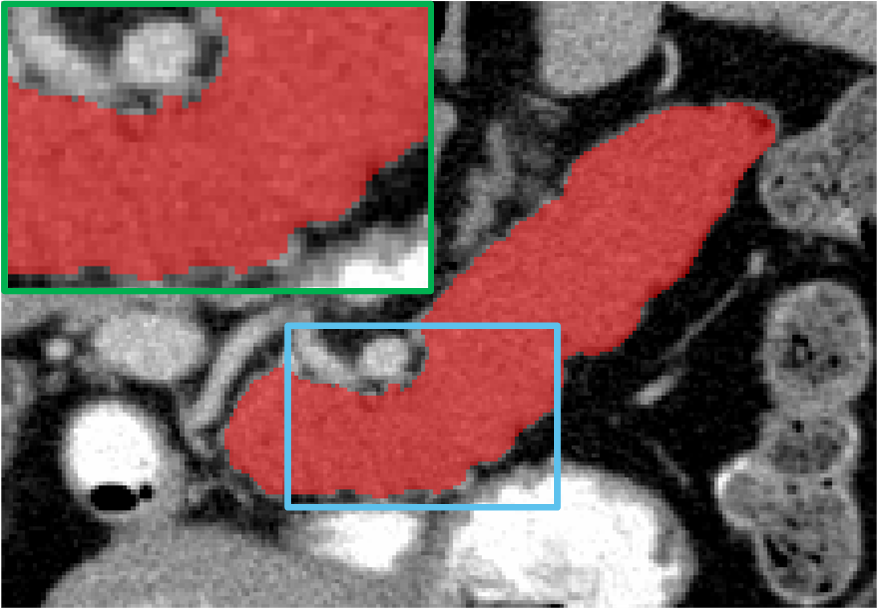}\vspace{5pt}
                \includegraphics[width=1\linewidth]{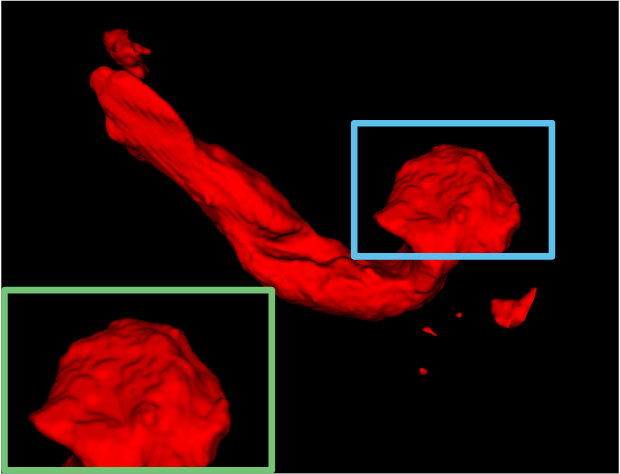} 
            \end{minipage}}
        	\centering
	\subfigure[Ours]{
            \begin{minipage}[b]{0.116\linewidth}
                \includegraphics[width=1\linewidth]{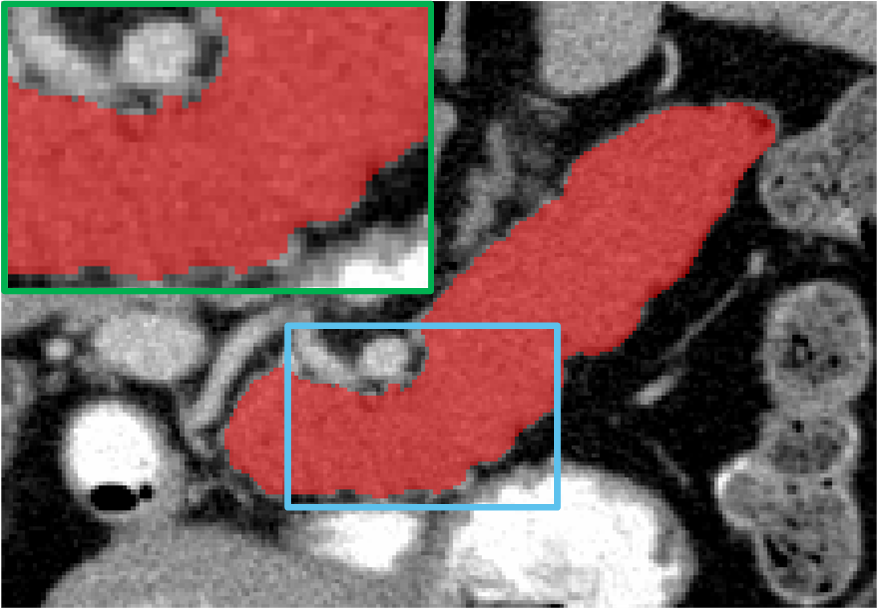}\vspace{5pt}
                \includegraphics[width=1\linewidth]{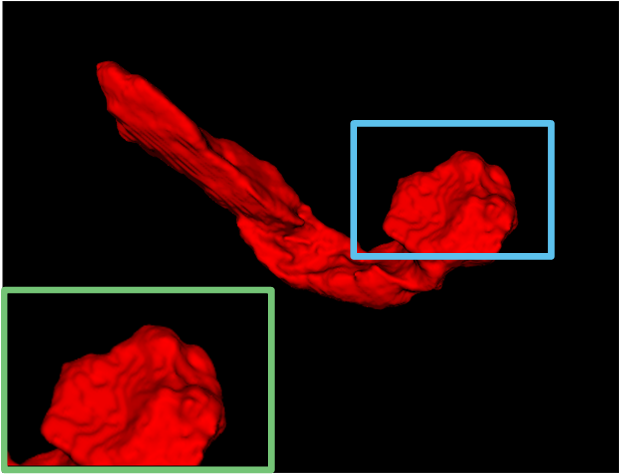} 
            \end{minipage}}
        \centering
	\subfigure[Ground Truth]{
            \begin{minipage}[b]{0.116\linewidth}
                \includegraphics[width=1\linewidth]{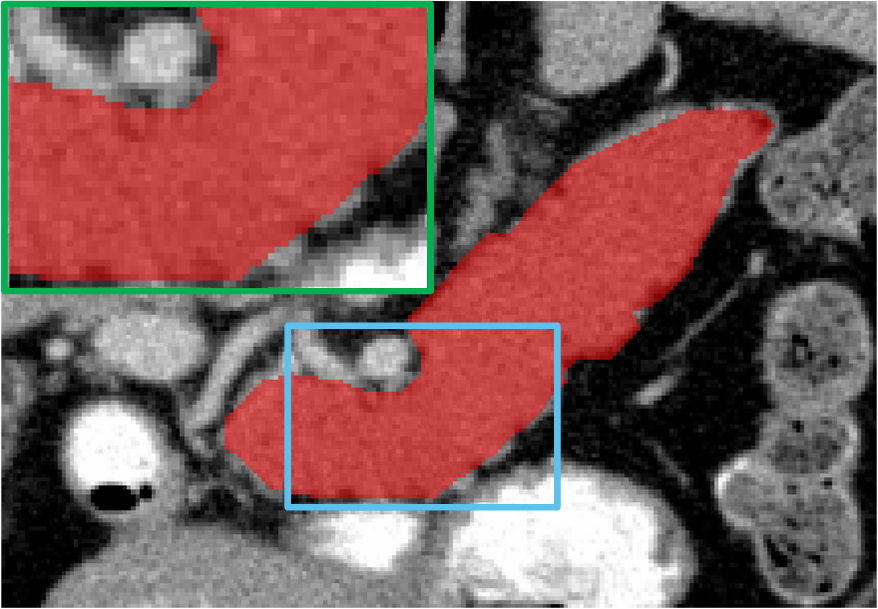}\vspace{5pt}
                \includegraphics[width=1\linewidth]{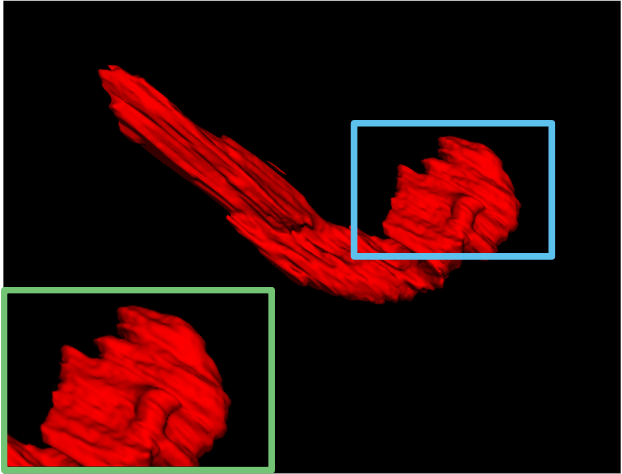} 
            \end{minipage}}
	\caption{Comparison of visual results on Pancreas-CT Dataset. The first row shows the results in 2D form, while the second row provides the visualizations in 3D form, where certain details have been enlarged for better clarity.}
	\label{figure5}
\end{figure*}

Yet within a notably challenging task, the proposed DCF demonstrates promising performance in both scenarios. Even with only 10\% of the data labeled, DCF significantly improves Dice scores from 55.10\% to 78.94\%, surpassing all other SSL methods. With 20\% labeled data, DCF achieves a Dice score of 81.65\%, outperforming again other cutting-edge competitors. It should be noted that our proposal exhibits the smallest disparity between 10\% and 20\% labeled scenarios, as illustrated in Figure \ref{figure_zzt}. The marginal 2.71\% variance in DCF's Dice scores between these proportions suggests its effective utilization of unlabeled data, thereby demonstrating robustness and generalization capabilities.
We further deliver visualization of the results obtained by DCF and others, as illustrated in Figure \ref{figure5}. It is observed that our results closely resemble the ground truth (GT) compared to those of other methods. Moreover, our method exhibits more precise boundary positioning and provides the more detailed information.

\vspace{-10pt}

\subsection{Result on ISIC}
\vspace{-2pt}
To further validate the generalizability of our proposed model, an additional verification is carried out using 2D images. Several state-of-the-art methods are re-implemented on the ISIC dataset, including MT \citep{tarvainen2017mean}, UA-MT \citep{yu2018pu}, CCT \citep{ouali2020semi}, CPS \citep{chen2021semi}, and UCMT \citep{Shen2023CotrainingWH}. The concerned results are presented in Table \ref{table3}. Similarly, two scenarios with 5\% and 10\% labeled data are respectively established. As given, we observe that DCF outperforms the other methods, demonstrating its superior generalization capability

\vspace{-10pt}

\section{Ablation Study}

Comprehensive ablation experiments are conducted on the final DCF architecture, validating the effectiveness of the tutoring mechanism and assessing the impact of various competitive approaches.

\textbf{Impact of Competitive treatments}: When evaluating student performance in the current iteration, we have multiple indicators to be chosen. In this experiment, we use Dice, CE, Jac, 95HD, and ASD, studying their individual and combined effects.

\begin{table}[htbp] 
\scriptsize
\centering
\caption{Ablations of different tutoring mechanisms on LA dataset.} 
\label{table5} 
\begin{tabular}{c|cccc}
\hline 
\multirow{2}{*}{ \textbf{Methods} }  & \multicolumn{4}{c}{\textbf{Metrics}} \\  
\cline{2-5}
& Dice(\%)$\uparrow$ & Jaccard(\%)$\uparrow$ & 95HD(voxel)$\downarrow$ & ASD(voxel)$\downarrow$  \\
\hline 
(1) &  $91.10$ & $83.72$ & $5.36$ & $1.50$ \\ 
(2) &  $90.26$ & $82.31$ & $5.68$ & $1.67$ \\ 
(3) &  $90.63$ & $82.92$ & $5.86$ & $1.60$ \\ 
(4) &  $90.70$ & $83.03$ & $5.68$ & $1.59$ \\ 
(5) &  $\textbf{91.44}$ & $\textbf{84.28}$ & $\textbf{5.24}$ & $\textbf{1.55}$ \\ 
\hline
\end{tabular}
\end{table}

When 20\% of labeled data is employed, Table \ref{table4} presents specific findings demonstrating that on the LA dataset, optimal outcomes are achieved with the utilization of Dice as the sole competitive metric. We believe that this is because the Dice metric outperforms other metrics in accurately evaluating model performance in the field of medical image segmentation. However, for varied tasks, it is also believed that alternative evaluation metrics should be contemplated.

\textbf{Effectiveness of Tutoring Mechanism}. To assess the engineered tutoring mechanism, various scenarios are devised: 1) Teachers refraining from tutoring. 2) Teachers tutoring irrespective of performance. 3) Alternating tutoring duties among students. 4) Teachers offering extra support to high-performing students. 5) Providing tutoring to low-performing students. The experimental outcomes on the LA dataset (20\% labeled data) are detailed in Table \ref{table5}.

It is illustrated that the model exhibits optimal performance when the teacher administers remediation to poorly performing students, thereby validating the efficacy of our remediation treatment. However, the efficacy of the model diminishes when remediation is provided exclusively to well-performing students or when simultaneous remediation is administered to two students. This decline may be due to an exacerbated variance between students. Regarding the randomized remediation method, we contend that its indiscriminate nature undermines its ability to yield favorable outcomes. 
\begin{table}[htbp] 
\scriptsize
\centering 
\caption{Variations in model performance on LA Dataset under differing evaluation schemes during student competition.} 
\label{table4}
\begin{tabular}{c|cccc}
\hline 
\multirow{2}{*}{ \textbf{Methods} }  & \multicolumn{4}{c}{\textbf{Metrics}} \\  
\cline{2-5}
& Dice(\%)$\uparrow$ & Jaccard(\%)$\uparrow$ & 95HD(voxel)$\downarrow$ & ASD(voxel)$\downarrow$  \\
\hline 
Dice &  $\textbf{91.44}$ & $\textbf{84.28}$ & $\textbf{5.24}$ & $\textbf{1.55}$ \\  
CE &  $90.50$ & $82.70$ & $6.02$ & $1.56$ \\  
Jac &  $89.32$ & $80.79$ & $7.43$ & $2.21$ \\   
ASD &  $90.80$ & $83.20$ & $5.51$ & $1.52$ \\  
95HD &  $90.72$ & $83.07$ & $5.75$ & $1.60$ \\   
Dice+Jac &  $89.07$ & $80.40$ & $6.56$ & $2.15$ \\  
Dice+CE &  $90.37$ & $82.51$ & $6.08$ & $1.58$ \\  
CE+Jac &  $89.44$ & $81.01$ & $7.66$ & $2.20$ \\ 
95HD+ASD &  $91.05$ & $83.62$ & $5.60$ & $1.48$ \\ 
CE+Jac+Dice &  $88.45$ & $79.43$ & $7.60$ & $2.39$ \\ 
\hline
\end{tabular}
\end{table}
In instances where no tutoring mechanism is employed, the model neither exacerbates variance among students nor achieves optimization, potentially yielding subpar results.

\vspace{-10pt}

\section{Conclusion}
\vspace{-2pt}
In this study, we present a novel semi-supervised framework for 3D medical image segmentation, which is specifically designed to tackle the challenge of tight coupling in the Teacher-Student structure within MT-based methods. In addition, a competitive tutoring mechanism is crafted to improve communication between models, thus mitigating the risk of model collapse resulting from the acquisition of erroneous knowledge. Besides, we employ weight distance and prediction distance to perform a detailed analysis of the state changes among the three networks throughout the training process. The effectiveness of our DCF in semi-supervised medical image segmentation is validated on three public benchmark datasets. Furthermore, we believe that our proposed DCF framework can serve as a plug-and-play solution, readily applicable across diverse SSL fields.
However, our method still suffers from certain limitations, \textit{e.g.}, the dynamic interplay between two students is mostly pronounced during the initial and middle periods, which would gradually wane as time progresses. Moving forward, we intend to further investigate strategies to enhance consistency between students. 

\vspace{-10pt}
\begin{ack}
This work was supported in part by the Key Program of Natural Science Foundation of Zhejiang Provinceunder under Grant LZ24F030012, and in part by the National Natural Science Foundation of China under Grant 62276232.
\end{ack}




\end{document}